\documentclass[conference]{IEEEtran}
\IEEEoverridecommandlockouts
\usepackage{cite}
\usepackage{amsmath,amssymb,amsfonts}
\usepackage{algorithmic}
\usepackage{microtype}
\usepackage{graphicx}
\usepackage{textcomp}
\usepackage{xcolor}
\usepackage{subfigure}
\usepackage{booktabs} 
\usepackage{hyperref}
\usepackage{url}
\usepackage{wrapfig}
\usepackage{graphicx}
\usepackage{multirow}
\usepackage{colortbl}
\usepackage{hhline}
\usepackage{tablefootnote}
\usepackage{resizegather}
\usepackage{cite}
\usepackage{tikz}
\usepackage{mathtools}
\usepackage{amssymb}

\def\BibTeX{{\rm B\kern-.05em{\sc i\kern-.025em b}\kern-.08em
    T\kern-.1667em\lower.7ex\hbox{E}\kern-.125emX}}
\begin{document}

\title{Instance-based Label Smoothing For Better Calibrated Classification Networks
\thanks{Paper has been accepted and will be published at ICMLA 2021.}
}

\author{\IEEEauthorblockN{Mohamed Maher}
\IEEEauthorblockA{\textit{Computer Science Department} \\
\textit{University of Tartu}\\
Tartu, Estonia \\
mohamed.abdelrahman@ut.ee}
\and
\IEEEauthorblockN{Meelis Kull}
\IEEEauthorblockA{\textit{Computer Science Department} \\
\textit{University of Tartu}\\
Tartu, Estonia \\
meelis.kull@ut.ee}
}

\maketitle

\begin{abstract}
Label smoothing is widely used in deep neural networks for multi-class classification. While it enhances model generalization and reduces overconfidence by aiming to lower the probability for the predicted class, it distorts the predicted probabilities of other classes resulting in poor class-wise calibration. Another method for enhancing model generalization is self-distillation where the predictions of a teacher network trained with one-hot labels are used as the target for training a student network. We take inspiration from both label smoothing and self-distillation and propose two novel instance-based label smoothing approaches, where a teacher network trained with hard one-hot labels is used to determine the amount of per class smoothness applied to each instance. The assigned smoothing factor is non-uniformly distributed along with the classes according to their similarity with the actual class. 
Our methods show better generalization and calibration over standard label smoothing on various deep neural architectures and image classification datasets.
\end{abstract}


\section{INTRODUCTION}

Recently, deep learning has been used in a wide range of applications. Thanks to the accelerated research progress in this field, deep \emph{neural networks} (DNNs) have been widely utilized in real-world multi-classification tasks including image, and text  classification. 
Although these classification networks are out-performing humans in many tasks, deploying them in critical domains that require making decisions which directly impact people's lives could be quite risky. Trusting the model's predictions with such decisions requires a high degree of confidence from a \emph{reliable} model.
Such model should be \emph{confidence-calibrated}, meaning that for example, the model's predictions at 99\% confidence are on average correct in 99\% of the cases \cite{guo2017calibration}.
However, on the 1\% of cases where the model is wrong, it often still matters which of the wrong classes is predicted, as some confusions can be particularly harmful (e.g. resulting in a wrong harmful medical treatment).
Therefore, the model should also be \emph{classwise-calibrated} \cite{kull2019beyond}, meaning that the predicted probabilities for each class separately should be calibrated, not only the top-1 predicted class as in confidence-calibration.

It has been shown that high capacity networks with deep and wide layers suffer from significant confidence calibration error \cite{guo2017calibration} as well as classwise calibration error \cite{kull2019beyond}.

\emph{Post-hoc calibration methods} such as temperature scaling \cite{guo2017calibration} or Dirichlet calibration \cite{kull2019beyond} learn a calibrating transformation on top of the trained model, to map the overconfident predicted probabilities into confidence- and classwise-calibrated predictions.

An alternative (or complementary) approach to post-hoc calibration is \emph{label smoothing} (LS) \cite{szegedy2016rethinking, muller2019does}
whereby the one-hot vector (hard) labels used for training a classifier are replaced by smoothed (soft) labels which are weighted averages of the one-hot vector and the constant uniform class distribution vector.
While originally proposed as a regularization approach to improve generalization ability \cite{szegedy2016rethinking}, LS has shown to make the neural network models robust against noisy labels \cite{lukasik2020does} and improve calibration also \cite{muller2019does}. 
In fact, a technique basically equivalent to LS for binary classification was already used to improve calibration nearly two decades earlier in the post-hoc calibration method known as logistic calibration or Platt scaling \cite{platt1999probabilistic}.

On the other hand, recent attempts to understand LS have shown some of its limitations. 
LS pushes the network towards treating all the incorrect classes as equally probable, while training with hard labels does not \cite{muller2019does,lienen2021label}.
This is in agreement with our experiments where models trained with LS tend to be good in confidence-calibration but not so good in classwise-calibration.
M{\"u}ller et al.~\cite{muller2019does} demonstrated that the smoothing mechanism leads to the loss of the similarity information among the dataset's classes.
LS was also found to impair the network performance in transfer learning and network distillation tasks \cite{muller2019does, kornblith2019better, maher2020}.

\paragraph{Outline and Contributions.} Section~\ref{sec:pre} introduces LS and methods to evaluate classifier calibration. 
In Section~\ref{sec:drawbacks}, we identify two limitations of LS: (1) LS is not varying the smoothing factor across instances and this is a missed opportunity, because different subsets of the training data benefit from different smoothing factors; (2) while the models trained with LS are quite well confidence-calibrated, they are relatively poorly classwise-calibrated.
In Section~\ref{sec:ils}, we propose two methods of instance-based LS called ILS1 and ILS2 to overcome the above 2 limitations, respectively.
Section~\ref{sec:prior} covers prior work, Section~\ref{sec:exp} compares the proposed methods against existing methods on real datasets, and Section~\ref{sec:conc} concludes our work.

\section{Preliminaries}
\label{sec:pre}

\subsection{Label Smoothing}
Consider a $K$-class classification task and an instance with features $\mathbf{x}$ and its (one-hot encoded) hard label vector $\mathbf{y}=(y_1, y_2,\dots,y_K)$. For each instance,
$y_j=1$ if $j=t$ where $t$ is the index of the true class out of the $K$ classes, and $y_j=0$ otherwise. 
We consider networks that are trained to minimize the cross-entropy loss which on the considered instance is $L=\sum_{j=1}^{K}(-y_j\log(\hat{p}_{j}))$, where $\mathbf{\hat{p}}=(\hat{p}_1,\dots,\hat{p}_K)$ is the model's predicted class probability vector.
Since $y_j$ is $0$ for all incorrect classes, the network is trained to maximize its confidence for the correct class, $L = -\log(\hat{p}_{t})$, while the probability distribution over all other incorrect classes is ignored. 

When training a network with LS, each (hard) one-hot label vector $\mathbf{y}$ is replaced with the smoothed (soft) label vector $\mathbf{y}^{LS}=(y^{LS}_1,\dots,y^{LS}_K)$ defined as $y^{LS}_{j} = y_{j} (1 - \epsilon) + \frac{\epsilon}{K}$ where $\epsilon$ is called \emph{the smoothing factor}.
For example, with 3 classes and smoothing factor $\epsilon=0.1$, the hard vector $(1,0,0)$ would get replaced by the soft vector $(1-\epsilon,0,0)+(\frac{\epsilon}{3}, \frac{\epsilon}{3}, \frac{\epsilon}{3})=(0.933, 0.033, 0.033)$.

\subsection{Evaluating Model Calibration}
Confidence is the highest probability in the classifier's prediction vector $\mathbf{\hat{p}}$.
The classifier is confidence-calibrated, if among all instances where the classifier has confidence $p$, the proportion of correct predictions is also $p$, for any $p\in[0,1]$ \cite{naeini2015obtaining}.
Confidence-calibration can be evaluated using the measure known as the confidence-ECE or simply ECE (Estimated Calibration Error as called by \cite{roelofs2020mitigating}, or also known as the Expected Calibration Error \cite{naeini2015obtaining}).
The possible confidences $[0,1]$ are binned into $N$ \emph{bins} $B_1,\dots,B_N$, typically with equal widths, i.e. $B_1 = [0, \frac{1}{N}]$ and $B_n = (\frac{n-1}{N}, \frac{n}{N}]$ for $n\in {2,3,...,N}$). 
The ECE (Equation \ref{eq:ece}) is calculated as the weighted average of the absolute difference between the mean model accuracy and mean confidence within each bin $B_n$ \cite{naeini2015obtaining}:
\begin{gather}
\label{eq:ece}
\mathsf{ECE} = \sum_{n=1}^N \left( \frac{|B_n|}{\sum_{i=1}^N |B_i|} \times |\mathsf{accuracy}(B_n) -\mathsf{confidence}(B_n)| \right).
\end{gather}
where $|B_n|$ is the number of instances in the bin $B_n$, and $\mathsf{accuracy}(B_n)$ is the proportion of correct predictions among these instances, and $\mathsf{confidence}(B_n)$ is the mean confidence.

The notion of classwise-calibration requires that for any class $j\in\{1,\dots,K\}$ and for any $p\in[0,1]$, the proportion of class $j$ among all instances with $\hat{p}_j=p$ is equal to $p$ \cite{kull2019beyond}. 
Classwise-calibration can be evaluated with the classwise-ECE (cwECE), defined as:
\begin{gather}
\label{eq:cwECE}
\mathsf{cwECE} = \frac{1}{K} \sum_{j=1}^K \sum_{n=1}^N
\left( \frac{|B_{n,j}|}{\sum_{i=1}^N |B_{i,j}|} \times |\mathsf{actual}(B_{n,j}) - \mathsf{predicted}(B_{n,j})| \right).
\end{gather}
where $|B_{n,j}|$ is the number of instances where the predicted probability $\hat{p}_j$ falls into the bin $B_n$, whereas $\mathsf{actual}(B_{n,j})$ and $\mathsf{predicted}(B_{n,j})$ are the  proportion of class $j$ and the average prediction $\hat{p}_j$ among those instances, respectively \cite{kull2019beyond}.
In our experiments, following Roelofs et al. \cite{roelofs2020mitigating}, we used 15 equal-width bins for both ECE and cwECE calculations.

\section{The Drawbacks of Standard Label Smoothing}
\label{sec:drawbacks}

Earlier work by 
M{\"u}ller et al.~\cite{muller2019does} has demonstrated that LS pushes the network towards treating all the incorrect classes as equally probable, also impairing the performance in network distillation tasks.
In the following, we study the drawbacks of LS in more detail on a synthetic classification task, where we know exactly what the ground truth Bayes-optimal class probabilities are.

\begin{figure}[!ht]
  \begin{center}
    \includegraphics[width=0.5\textwidth]{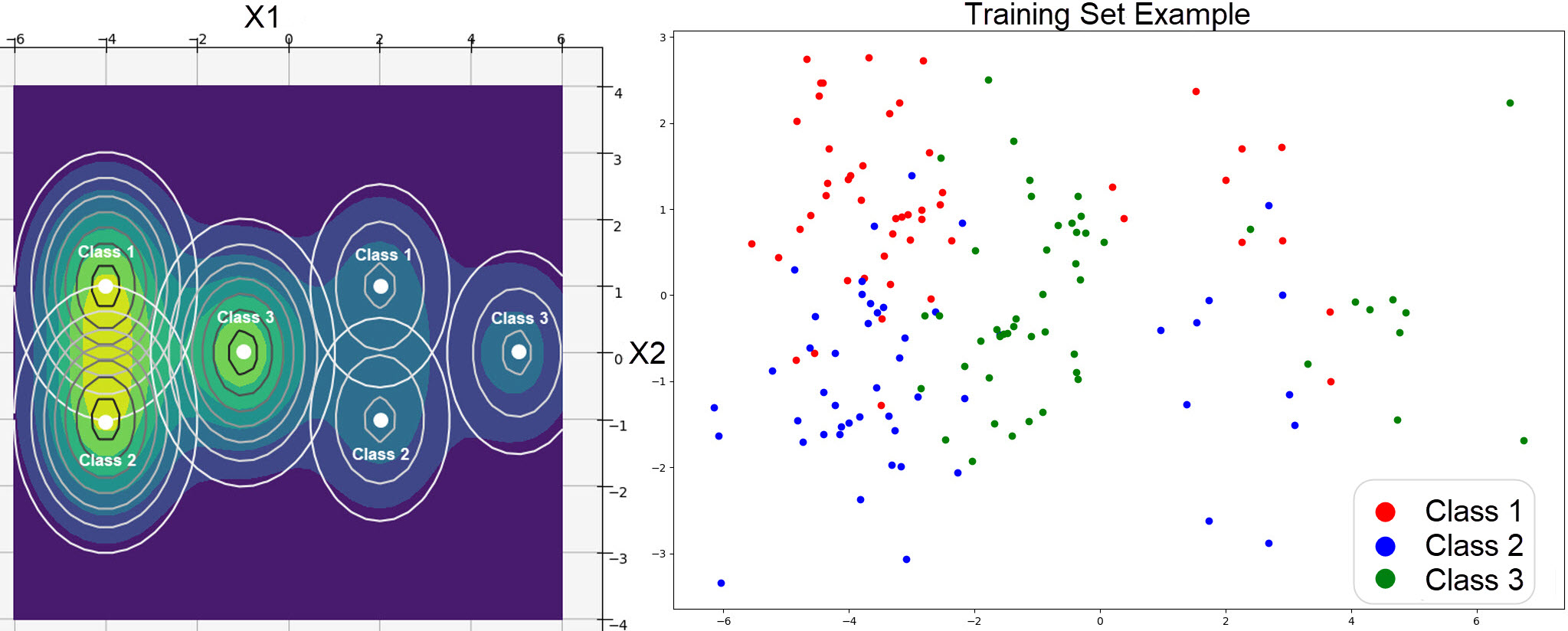}
  \end{center}
  \caption{Left:Heat-map of the total density function of the generative model of 3-classes synthetic dataset. Right:An example of a training set sampled from the generative model of the synthetic dataset.}
  \label{fig:synthetic-dataset}
\end{figure}

\paragraph{Experimental setup.} The synthetic task is shown in Figure \ref{fig:synthetic-dataset} and consists of 2 features (x- and y-axis), 3 classes (1, 2, 3), each as a mixture of 2 bivariate Gaussians, with 80\% of mass in the `left' or `dense' Gaussian, and 20\% in the `right' or `sparse' Gaussian. 
The means of class 1 Gaussians are $(-4,1)$ and $(2,1)$; for class 2 Gaussians $(-4,-1)$ and $(2,-1)$; and for class 3 Gaussians $(-1,0)$ and $(5,0)$.
The covariance matrix is the identity matrix for all the 6 Gaussians.

Due to the chosen constellation, most instances belonging to class $1$ are more probable to belong to class $2$ than to class $3$. 
Such asymmetry makes it important for the models to have non-equal probabilities for the wrong classes, in order to be classwise-calibrated.
The dense and sparse parts have been created to test whether the same amount of LS is appropriate in both parts.

\paragraph{Results that motivate ILS2.} Using the above generative model, we sample $100$ replicate datasets using different random seeds, each with training, validation, and testing sets of sizes $50$, $50$, and $5000$ instances per class, respectively. 
I.e., the training set has $150$ instances, with about $40$ instances generated from the dense and $10$ instances from the sparse Gaussian of each class. 
The training and validation sets were chosen to be so small to have enough overfitting.
As the task is simple, a large training set would lead to near-optimal models.
We trained feed-forward neural networks (5 hidden layers, 64 units per layer, ReLU activation) on each dataset, minimizing the cross-entropy loss, without LS (referred to as: No-LS) and with LS (referred to as: Standard LS).
When training with LS, we used the smoothing factor from $\epsilon\in\{0.001, 0.005, 0.01, 0.03, 0.05, ..., 0.19\}$ that resulted in the best validation loss. 
To test the effect of using large smoothing factors we included the version with a fixed smoothing factor of $0.2$ into the comparison (referred to as: Standard LS-0.2).
To evaluate whether there is additional benefit from post-hoc calibration, we optionally applied temperature scaling \cite{guo2017calibration}, where the temperature parameter $T$ was optimized on the validation set (finding the best combination of $T$ and $\epsilon$).

\begin{table*}[!ht]
\centering
\caption{Comparison between training with/out LS. Acc Dense/Sparse represent the accuracy on the test set sampled from the dense/sparse regions separately. LS implicitly calibrates over-confident predictions yet distorts the classes of lower probability predictions.}
\label{tab:motivate-calibration}
\resizebox{0.9\textwidth}{!}{
\begin{tabular}{|c|c|l|c|c|c|c|c|c|c|c|c|} 
\hline
\begin{tabular}[c]{@{}c@{}} \textbf{Training}\\\textbf{Set} \end{tabular} & \begin{tabular}[c]{@{}c@{}} \textbf{Validation}\\\textbf{Set} \end{tabular} & \multicolumn{1}{c|}{\textbf{Method} } & \textbf{TempS}  & \textbf{Acc} & \begin{tabular}[c]{@{}c@{}}\textbf{Acc}\\\textbf{Dense} \end{tabular} & \begin{tabular}[c]{@{}c@{}}\textbf{Acc}\\\textbf{Sparse} \end{tabular} & \begin{tabular}[c]{@{}c@{}}\textbf{Cross}\\\textbf{Entropy}\end{tabular} & \textbf{ECE}  & \textbf{cwECE}  & \begin{tabular}[c]{@{}c@{}}\textbf{Avg}\\\textbf{Temp}\end{tabular} & \textbf{Avg~$\epsilon$}  \\ 
\hline
\multirow{9}{*}{\begin{tabular}[c]{@{}c@{}}All \\ (both Dense \\ and Sparse) \end{tabular}} & \multirow{7}{*}{All} & \textit{Bayes-Optimal }  & No & \textit{81.98\% }  & \textit{82.94}\% & \textit{78.14}\% & \textit{0.4444}  & \textit{0.0067}  & \textit{0.0234}  & \textit{1.000} &  \\ 
\cline{3-11}
 &  & No LS & No & 78.45\% & 81.43\% & 66.54\% & 0.5589 & 0.0426 & 0.0668 & 1.000 &  \\ 
\cline{3-11}
 &  & No LS & Yes & 78.45\% & 81.43\% & 66.54\% & 0.5446 & 0.0311 & \textbf{0.0632}  & 1.202 &  \\ 
\cline{3-12}
 &  & Standard LS & No & 79.44\%  & 81.17\% & 72.51\% & 0.5318 & 0.0293 & 0.0723 & 1.000 & 0.061 \\ 
\cline{3-12}
 &  & Standard LS & Yes & \textbf{79.48}\% & 81.24\% & 72.44\% & \textbf{0.5238}  & 0.0263 & 0.0701 & 0.946 & 0.079 \\ 
\cline{3-12}
 &  & Standard LS-0.2 & No & 79.39\% & 81.11\% & 72.53\% & 0.5439 & 0.0650 & 0.0785 & 1.000 & 0.200 \\ 
\cline{3-12}
 &  & Standard LS-0.2 & Yes & 79.39\% & 81.11\% & 72.53\% & 0.5301 & \textbf{0.0241}  & 0.0707 & 0.804 & 0.200 \\ 
\hhline{|~-----------|}
 & {\cellcolor[rgb]{0.922,0.922,0.922}}Dense & {\cellcolor[rgb]{0.922,0.922,0.922}}Standard LS & {\cellcolor[rgb]{0.922,0.922,0.922}}No & {\cellcolor[rgb]{0.922,0.922,0.922}}79.06\% & {\cellcolor[rgb]{0.922,0.922,0.922}}\textbf{81.53\%}  & {\cellcolor[rgb]{0.922,0.922,0.922}}69.18\% & {\cellcolor[rgb]{0.922,0.922,0.922}}0.5326 & {\cellcolor[rgb]{0.922,0.922,0.922}}0.0287 & {\cellcolor[rgb]{0.922,0.922,0.922}}0.0749 & {\cellcolor[rgb]{0.922,0.922,0.922}}1.000 & {\cellcolor[rgb]{0.922,0.922,0.922}}0.047 \\ 
\hhline{|~-----------|}
 & {\cellcolor[rgb]{0.922,0.922,0.922}}Sparse & {\cellcolor[rgb]{0.922,0.922,0.922}}Standard LS & {\cellcolor[rgb]{0.922,0.922,0.922}}No & {\cellcolor[rgb]{0.922,0.922,0.922}}78.73\% & {\cellcolor[rgb]{0.922,0.922,0.922}}81.05\% & {\cellcolor[rgb]{0.922,0.922,0.922}}\textbf{74.47\%}  & {\cellcolor[rgb]{0.922,0.922,0.922}}0.5365 & {\cellcolor[rgb]{0.922,0.922,0.922}}0.0375~ & {\cellcolor[rgb]{0.922,0.922,0.922}}0.0778 & {\cellcolor[rgb]{0.922,0.922,0.922}}1.000 & {\cellcolor[rgb]{0.922,0.922,0.922}}0.122 \\
\hline
\end{tabular}}
\end{table*}

Table~\ref{tab:motivate-calibration} shows the results of experiments on the synthetic task, evaluated with accuracy, cross-entropy, (confidence-)ECE and classwise-ECE.
As expected, for each method (No LS, Standard LS, Standard LS-0.2), applying post-hoc temperature scaling improves cross-entropy, ECE and classwise-ECE, while keeping the accuracy the same (except for Standard LS, where the difference is due the combined tuning of $T$ and $\epsilon$).
Standard LS is improving accuracy, and also improving cross-entropy and ECE more than temperature scaling alone (with No-LS); this is in agreement with the results of M{\"u}ller et al.~\cite{muller2019does} and can be explained by LS reducing overconfidence.
However, Standard LS has worse performance in terms of classwise-ECE.
Thus, while LS is beneficial for confidence calibration, it can be harmful for classwise calibration.
This could be explained by the definition of LS: it is introducing the same non-zero amount of smoothing towards all wrong classes, while actually, some wrong classes are closer and some further away from the instance.
Even temperature scaling is not able to undo the harm of LS, as even then the classwise-ECE remains worse than for the model trained without LS.
Hence, the model trained without LS contains more information about the similarity of classes, as encoded by the predicted class probabilities.
This motivates our work to propose ILS2 which distributes the smoothing factor unevenly across the wrong classes.

\paragraph{Results that motivate ILS1.} In order to investigate the effect of using the same smoothing factor along all the instances, we repeated the experiment for Standard LS two more times, but now tuning $\epsilon$ to give the lowest cross-entropy on a reduced validation set: either only the dense (left) part of the dataset, or the sparse (right) part. 
As expected, Table~\ref{tab:motivate-calibration} (rows: Dense, Sparse; columns: Acc Dense, Acc Sparse) shows that the model with $\epsilon$ tuned on the dense part has slightly higher accuracy on the dense part (81.53\% vs 81.05\%), and the one tuned on the sparse part is more accurate on the sparse part (74.47\% vs 69.18\%).
However, importantly, the learned smoothing factor $\epsilon$ is radically different for these two models.
Averaged over the 100 replicate experiments, the best $\epsilon$ for the dense part is much lower ($0.047$) than on the sparse part ($0.122$).
Since the model is subject to more overfitting in the sparse region, a higher smoothing factor $\epsilon$ is preferred to reduce the model's over-confidence.
This motivates our work to propose ILS1, varying the smoothing factor across instances.

\begin{figure*}
  \begin{center}
    \includegraphics[width=\textwidth]{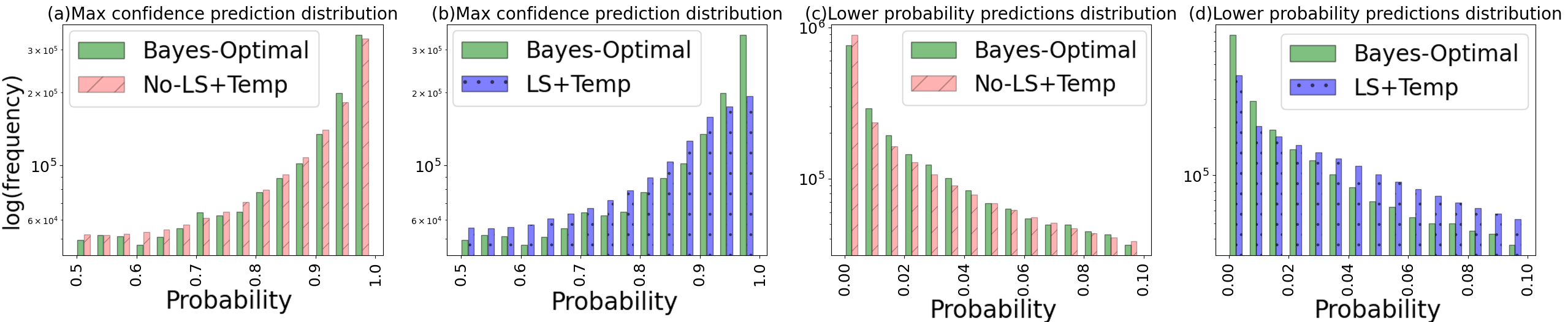}
  \end{center}
  \caption{Comparisons of the histogram distributions of the probability predictions between the Bayes-Optimal model and training with/without LS and with temperature scaling.}
  \label{fig:prediction-diversity}
\end{figure*}

\paragraph{LS in network distillation.}
Another drawback was identified by \cite{muller2019does}, observing that LS impairs \emph{network distillation} - a process where a `student' network is trained to mimic the outputs of a `teacher' network which is trained on the original data \cite{hinton2015distilling}.
They showed that the student network performed worse when training the teacher network with LS as opposed to without LS.
We tested it on our synthetic data as well, using the identical architecture (the same as described above) both for the student and the teacher. 
This is known as self-distillation and has been shown to have a regularization behavior for the trained models \cite{zhang2019your, mobahi2020self,yang2019training}.
Results are shown in Table~\ref{tab:motivate-distillation}, confirming worse performance of the student in all measures (accuracy, cross-entropy, ECE, cwECE) when the teacher is trained with LS compared to without LS.
As we take inspiration from distillation in proposing our method ILS1, this drawback of LS is important to keep in mind.

\paragraph{Distribution of predicted probabilities.} 
Two out of three of the above identified drawbacks relate to the distribution of predicted probabilities. 
First, we showed that LS biases the predicted probabilities of the wrong classes, and last, we observed that LS impairs distillation as the teacher's predicted probabilities are not as informative as without LS.
Therefore, we study the distribution of predicted probabilities when the model is trained on the synthetic dataset with or without LS.

Figure~\ref{fig:prediction-diversity}(a) compares the distribution of confidence (the highest probability among the 3 predicted class probabilities) for the Bayes-optimal model and the model trained without LS. 
There are some differences (note the logarithmic y-axis!) but overall the distributions are quite similar.
However, when comparing the Bayes-optimal model with the model trained with LS (with temperature scaling), there are quite clear differences as seen in Figure~\ref{fig:prediction-diversity}(b): even after calibration with temperature scaling the model reaches top confidence on much fewer instances and instead has low confidence on more instances.

Similar is true for the other classes (the two non-highest probabilities among the 3 predicted class probabilities) as seen in Figures~\ref{fig:prediction-diversity}(c,d).
Indeed, the match between Bayes-optimal model and the model trained without LS is quite good, whereas the model trained with LS is outputting extreme predictions (the bin nearest to 0) much less frequently than the Bayes-optimal model.
These results help to explain the results in Table~\ref{tab:motivate-calibration} where cwECE is worse for Standard LS with temperature scaling than for No LS with temperature scaling.
This motivates the use of No LS with temperature scaling in our proposed instance-based label smoothing method ILS1, which we will discuss next.

\begin{table}[!ht]
\centering
\caption{Comparison between using teacher trained with/without LS in self-distillation.}
\label{tab:motivate-distillation}
\resizebox{0.4\textwidth}{!}{
\begin{tabular}{|l|c|c|c|c|} 
\cline{2-5}
\multicolumn{1}{l|}{} & \multicolumn{4}{c|}{\textbf{Student Performance}} \\ 
\hline
\multicolumn{1}{|c|}{ \textbf{Teacher} } & \textbf{Accuracy}  & \textbf{CrossEntropy}  & \textbf{ECE}  & \textbf{cwECE}  \\ 
\hline
\textit{Bayes-Optimal}  & \textit{81.92\% }  & \textit{0.4451}  & \textit{0.0137}  & \textit{0.0338}  \\ 
\hline
No LS & \textbf{78.47\% } & \textbf{0.5592}  & \textbf{0.0405}  & \textbf{0.0649}  \\ 
\hline
Standard LS & 78.43\% & 0.5621 & 0.0408 & 0.0726 \\
\hline
\end{tabular}}
\end{table}

\section{Instance-based Label Smoothing}
\label{sec:ils}
Based on the above drawbacks of standard LS, we propose two instance-based label smoothing (ILS) methods ILS1 and ILS2, where in ILS1 the instances have different smoothing factors; and in ILS2 the instances have the same smoothing factor but it is distributed non-evenly between the incorrect classes. 
Furthermore, we propose the combination of these methods that we call ILS, which combines the methods ILS1 and ILS2.

\begin{figure}
  \begin{center}
    \includegraphics[width=0.4\textwidth]{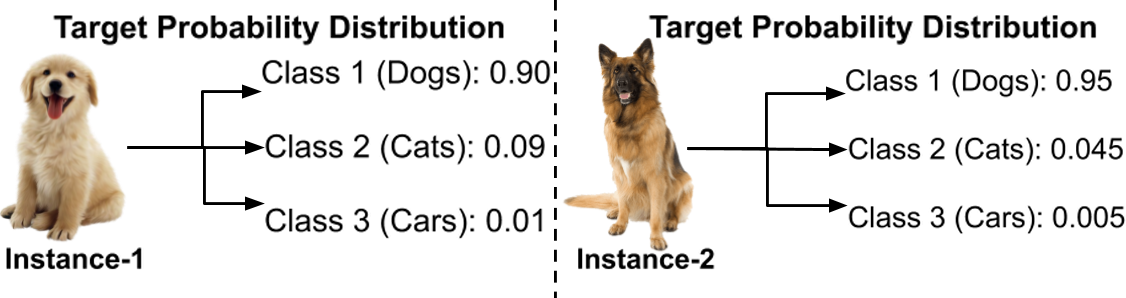}
  \end{center}
  \caption{An example of the target probability distribution of two dog instances. The cat class has a higher target probability than a car class. The left instance is more similar in shape to cats; So, a higher smoothing factor is used.}
  \label{fig:example}
\end{figure}

As an illustration of our proposal, imagine an image dataset with three different classes: dogs, cats and cars. 
As cats and dogs are more similar to each other than to cars, a dog instance should have a higher target probability to be a cat than a car. Similarly, a big-sized dog should have higher target confidence (hence, smaller smoothing factor) than a small dog that is close in shape to cats, see Figure~\ref{fig:example}.

\subsection{ILS1: Varying the Smoothing Factor Magnitude}
\label{sec:ils1}
In the above experiments on the synthetic dataset (Figure \ref{fig:synthetic-dataset}), we observed that the optimal smoothing factor was very different for the dense and sparse parts of the instance space. 
However, developing a method to choose the smoothing factor based on the density is problematic: (1) density estimation in high-dimensional spaces such as images is very hard; and (2) the amount of overlap between the classes is probably another important factor in addition to the density.
Instead, we opted for a simpler and more general approach to first train a network without LS to reveal which instances are easier or harder (higher or lower probability for the true class) and use this information to determine the instance-specific smoothing factors when training the final model (with the same architecture as the first network). 
Hence, in ILS1, we are taking inspiration from self-distillation \cite{zhang2019your} in the sense of first training a teacher, but differently from self-distillation we only use the teacher to determine the smoothing factor for the student, while still using the original (but smoothed) labels to train the student.
Due to the demonstrated drawbacks of LS in self-distillation, we train the teacher without LS and with temperature scaling.

\begin{figure}
  \begin{center}
    \includegraphics[width=0.35\textwidth]{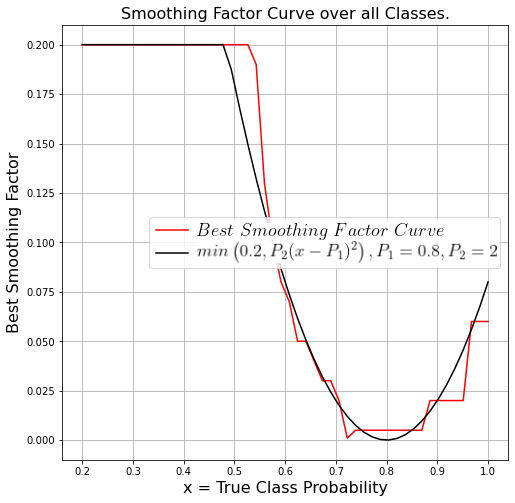}
  \end{center}
  \caption{The best smoothing factors corresponding to each group of instance with the same range of predicted true class confidence by the Bayes-Optimal model.}
  \label{fig:ils1}
\end{figure}

The key question is how to map the teacher's predicted true class probability $\hat{p}^{\mathsf{teacher}}_t$ into a smoothing factor to be used on this instance when training the student.
Instead of describing our proposed method ILS1 immediately, we first describe our reasoning towards this method.
We started by addressing the search for mappings from teacher's predictions to student's smoothing factors empirically on the synthetic dataset.
Running a separate experiment with each possible map would be intractable and thus we first just trained models with Standard LS (same $\epsilon$ for all instances) using different smoothing factors $\epsilon$ ranging from $0.001$ to $0.2$. 
Based on the predictions of all these (LS-student) models we addressed the question of which one of these is the best on the instances $\mathcal{I}_{\hat{p}^{\mathsf{teacher}}_t}$ where the teacher's predicted true class probability is $\hat{p}^{\mathsf{teacher}}_t$.
Here we came up with a naive hypothesis that the best LS-student is the one which has approximately the same average predicted true class probability $\hat{p}^{\mathsf{teacher}}_t$ on these instances.
In other words, we hypothesized that the instances $\mathcal{I}_{\hat{p}^{\mathsf{teacher}}_t}$ should have the smoothing factor $\epsilon$ such that when training a model on all instances with LS using $\epsilon$, then this model's average predicted true class probability among instances $\mathcal{I}_{\hat{p}^{\mathsf{teacher}}_t}$ would be approximately equal to $\hat{p}^{\mathsf{teacher}}_t$.
The red curve in Figure~\ref{fig:ils1} shows the best values for $\epsilon$ for each $\hat{p}^{\mathsf{teacher}}_t$ obtained this way, when using the Bayes-optimal model as the teacher.
Note that in practice, we define 50 groups of instances $\mathcal{I}_{\hat{p}^{\mathsf{teacher}}_t}$ by binning the values of $\hat{p}^{\mathsf{teacher}}_t$ into 50 bins from 0.2 to 1.0 (discarding instances below 0.2 as they rarely have any instances).
For comparison, somewhat similar curves are obtained when using the model trained without LS and with temperature scaling as the teacher, see Appendix\footnote{https://github.com/mmaher22/Instance-based-label-smoothing/blob/master/Appendix.pdf\label{appendix_url}} Figures 2,3. 
Furthermore, perhaps surprisingly, similar shapes are obtained also on real datasets with different network architectures and different numbers of classes, see Appendix\textsuperscript{\ref{appendix_url}} Figure \ref{fig:eg-real}.

This observation motivated the approach that we have taken with ILS1: choose a parametric family of functions which can approximate these \emph{best smoothing factor curves} and tune the parameters of this family on the validation set to minimize cross-entropy. 
That is, we try out some small number (30 in our experiments) of smoothing factor curves from this family; for each curve train a student model which uses this curve to determine the smoothing factor in each group $\mathcal{I}_{\hat{p}^{\mathsf{teacher}}_t}$ of instances; choose the student that is best on the validation data. 
There is a lot of freedom in choosing the family of functions and our ad hoc choice was to use parabolas which touch the x-axis and are clipped to the range $[0,0.2]$ (as we did not study smoothing factors higher than $0.2$):
\begin{equation}
\label{eq:ils1}
    \begin{array}{l}
        \epsilon^{ILS1}_x = min\left( 0.2, P_2(\hat{p}^{\mathsf{teacher}}_{t}(x)-P_1)^2  \right).
    \end{array}
\end{equation}
where $P_1,P_2\in\mathbb{R}$ are the two parameters.
In the experiments, we have used the ranges $[1,6]$ and $[0.8,0.99]$ for $P_1$ and $P_2$, respectively, as these ranges are sufficient to approximate the shapes that we have seen on the real data, see the Appendix B\textsuperscript{\ref{appendix_url}}. 
The selected smoothing factor is used the same way as in standard LS: the network is trained to minimize the cross-entropy loss with the new smoothed target probability distribution (i.e. $y^{ILS1}_{j} = y_{j} (1 - \epsilon^{ILS1}_x) + \frac{\epsilon^{ILS1}_x}{K}$). We have additionally experimented with a sinusoidal family $0.1(sin(p_2(x - p_1))+1)$ where $p_2$, $p_1$ are tuned similar to ILS1. On the synthetic data, this resulted in slightly worse performance than the quadratic family but still outperforming Standard LS (See the Appendix\textsuperscript{\ref{appendix_url}}). Consideration of alternative families and finding theoretical justifications for the families remains as future work.

We applied ILS1 on the synthetic data and its performance was better than Standard LS for (accuracy, cross-entropy, ECE and cwECE all improved).
We then took its tuned parameters $P_1=0.8$ and $P_2=2.0$ and plotted the respective smoothing factor curve, see the black curve in Figure~\ref{fig:ils1}.
Its very close fit with the best smoothing factor curve (shown in red) can be considered surprising because the black curve was not learned by fitting the red!
While the red curve was obtained through experiments with Standard LS with different values of $\epsilon$, the black curve only used ILS1 with instance-based smoothing and then tuning on the validation set.
Our interpretation is that our naive hypothesis was close enough to the truth in this case.

\subsection{ILS2: Uneven smoothing between incorrect classes}
\label{sec:ils2}
As noted in Section~\ref{sec:drawbacks}, LS is introducing the same $\epsilon/K$ amount of bias towards all incorrect classes, while actually, some wrong classes are closer and some other are further away from the instance.

Here we also use the inspiration from distillation and use the teacher network to guide the method that we call ILS2. As the teacher (NoLS+TempS) was the best class-wise calibrated model in Table \ref{tab:motivate-calibration}, we want to trust it when it says that an instance is for example twice as likely to belong to class i than class j. This can only be achieved within ILS2 if the mapping is linear in all classes (except the actual class).
However, since the bias is about the incorrect classes, we have decided to only use the teacher to inform the incorrect class probabilities while for the correct class probability we use the same value $1-\epsilon$ regardless of what the teacher's predicted probability was for that class:
\begin{equation}
\label{eq:ils2}
y^{ILS2}_{j} =
     \begin{cases}
       \frac{\epsilon \cdot \hat{p}_{j}^{teacher}(x)}{1 - \hat{p}_{t}^{teacher}(x)} &\quad\text{if } j \neq t \\
       1 - \epsilon &\quad\text{if } j = t.
     \end{cases}
\end{equation}
where $1 - \hat{p}_{t}^{teacher}(x)$ is the normalizing constant that is needed to ensure that the probabilities add up to 1.
Following the same approach as proposed by Hinton et al.~\cite{hinton2015distilling}, we tune the teacher's temperature parameter to minimize not the teacher's cross-entropy but instead the student's cross-entropy.
This means we train several students with different temperature values in the teacher, in our experiments we use $T\in\{1, 2, 4, 8, 16\}$.

\subsection{ILS: Combining ILS1 and ILS2}

To address all the identified drawbacks of LS simultaneously, we propose to combine ILS1 and ILS2 into a method that we refer to as ILS.
ILS trains a teacher network without LS and with temperature scaling, and then trains many students with instance-based LS with different values of $P_1$ and $P_2$ for ILS1 and T for ILS2, where the smoothing factor $\epsilon$ in Eq. (\ref{eq:ils2}) is replaced with $\epsilon^{ILS1}_x$ from Eq. (\ref{eq:ils1}).
The best student is selected based on cross-entropy on the validation set.

\section{Prior Work}
\label{sec:prior}
Liu and JaJa~\cite{liu2020class} proposed a class-based LS approach that assigns a target probability distribution corresponding to the similarity between the classes with the actual target class. Several semantic similarity measurements were tried out like calculating the Euclidean distances of raw inputs or the latent encoding produced from an auto-encoder model \cite{berthelot2018understanding}. Similarly, Yun et al.~\cite{yun2020regularizing} introduced a class-wise regularization method that penalizes the KLD of predictive distributions among instances belonging to the same class. During training, self-distillation was applied on wrong predictions using the output of correctly classified instances within the same class. This method was found to reduce overconfidence and intra-class variations while improving the network generalization.
Kim et al.~\cite{kim2020self} proposed self-distillation which softens the hard target labels during training by progressively distilling the own knowledge of a network. First, a teacher is trained with hard targets followed by training a student of the same architecture by minimizing the divergence between its own and the teacher predictions. Zhang and Sabuncu~\cite{zhang2020self} showed empirically that the improved generalization using distillation is correlated with an increase in the diversity of the predicted confidences. Following this, the authors relate self-distillation to standard LS. Finally, they propose an randomly-smoothed instance-based LS approach where different smoothing factors are randomly sampled out of a beta-distribution and assigned to each instance based on the prediction uncertainty, to encourage the model for more diverse confidence levels predictions. In \cite{lienen2021label}, Lienen and H{\"u}llermeier proposed a generalization to standard LS named label relaxation. The learner chooses one from a set of target distributions instead of using the precise target distribution of standard LS that could incorporate an undesirable bias during training. Reed et al.~\cite{reed2014training} proposed a soft bootstrapping method (BS\_Soft), where the network is trained using continuously updated soft targets of a convex combination of the training labels and the current network predictions. The resulting objective is equivalent to a softmax regression with minimum entropy regularization, which encourages the network to have a high confidence in predicting labels even for the unlabeled instances. Bootstrapping was found to improve the network accuracy in semi-supervised multi-class tasks, and implicitly reduce the overconfidence for noisy labels; however, it does not care about the class-wise calibration performance.

Temperature scaling was shown as the most effective calibration method in deep neural networks \cite{guo2017calibration}. However, it does not improve network generalization. M{\"u}ller et al.~\cite{muller2019does} showed empirically that LS improved the network generalization and could implicitly calibrate its maximum confidence predictions. On the other hand, it degrades the class similarity information learned by the network, which affects the transfer learning and distillation performance negatively. Mix-up regularization trains the classifier networks with augmented data instances using linear interpolations of a pair of the instances' features along with their labels leading to generalization improvement \cite{zhang2017mixup} but it hurts the network calibration \cite{liu2020class}.

\begin{table*}
\centering
\caption{The performance in terms of cross-entropy loss, accuracy, ECE, and cwECE on the testing set. Best results are marked in \textbf{bold}. Temp denotes temperature scaling. The lower average ranking is better. Subscripts represent the ranking of the method.}
\label{tab:results}
\resizebox{0.84\textwidth}{!}{\begin{tabular}{|l|c|cccccccc|cccc|} 
\hline
\textbf{Model} & \multicolumn{1}{l|}{\textbf{Dataset}} & \textbf{No LS} & \begin{tabular}[c]{@{}c@{}}\textbf{No LS+}\\\textbf{TempS}\end{tabular} & \textbf{Dis} & \textbf{cLS} & \textbf{BS\_Soft} & \textbf{Beta} & \textbf{LS} & \begin{tabular}[c]{@{}c@{}}\textbf{LS+}\\\textbf{TempS}\end{tabular} & \textbf{ILS1} & \textbf{ILS2} & \textbf{ILS} & \begin{tabular}[c]{@{}c@{}}\textbf{ILS+}\\\textbf{TempS}\end{tabular} \\ 
\hline
\multicolumn{14}{|c|}{\textbf{\textbf{Cross-Entropy}}} \\ 
\hline
LeNet & \multirow{5}{*}{CIFAR10} & $0.858_{12}$ & $0.858_{11}$ & $0.854_{10}$ & $0.811_{8}$ & $0.833_{9}$ & $0.787_{5}$ & $0.808_{7}$ & $0.800_{6}$ & $0.723_{3}$ & $0.784_{4}$ & $0.720_{2}$ & \textbf{$0.706_{1}$} \\
ResNet110 &  & $0.312_{11}$ & $0.265_{2}$ & $0.311_{10}$ & $0.284_{5}$ & $0.309_{9}$ & $0.362_{12}$ & $0.308_{8}$ & $0.293_{6}$ & $0.293_{7}$ & $0.272_{4}$ & $0.270_{3}$ & \textbf{$0.264_{1}$} \\
ResNet110SD &  & $0.310_{9}$ & $0.274_{4}$ & $0.314_{10}$ & $0.289_{6}$ & $0.318_{11}$ & $0.420_{12}$ & $0.303_{8}$ & $0.293_{7}$ & $0.289_{5}$ & $0.269_{3}$ & $0.266_{2}$ & \textbf{$0.226_{1}$} \\
DenseNet40 &  & $0.331_{11}$ & \textbf{$0.294_{1}$} & $0.301_{3}$ & $0.323_{10}$ & $0.311_{9}$ & $0.334_{12}$ & $0.306_{6}$ & $0.299_{2}$ & $0.311_{8}$ & $0.301_{4}$ & $0.307_{7}$ & $0.306_{5}$ \\
ResNet32W &  & $0.206_{11}$ & $0.154_{3}$ & $0.168_{8}$ & $0.170_{9}$ & $0.163_{6}$ & $0.240_{12}$ & $0.172_{10}$ & $0.159_{4}$ & $0.164_{7}$ & $0.162_{5}$ & $0.153_{2}$ & \textbf{$0.149_{1}$} \\ 
\hline
LeNet & \multirow{5}{*}{CIFAR100} & $2.361_{9}$ & $2.350_{8}$ & $2.704_{12}$ & $2.221_{4}$ & $2.540_{11}$ & $2.452_{10}$ & $2.292_{7}$ & $2.288_{6}$ & $2.183_{3}$ & $2.274_{5}$ & $2.173_{2}$ & \textbf{$2.132_{1}$} \\
ResNet110 &  & $1.382_{12}$ & $1.183_{4}$ & $1.163_{3}$ & $1.218_{6}$ & $1.292_{10}$ & $1.298_{11}$ & $1.283_{9}$ & $1.248_{8}$ & $1.222_{7}$ & $1.186_{5}$ & $1.103_{2}$ & \textbf{$1.055_{1}$} \\
ResNet110SD &  & $1.070_{11}$ & $0.992_{7}$ & $1.065_{10}$ & $0.954_{5}$ & $1.059{9}$ & $1.165_{12}$ & $0.993_{8}$ & $0.987_{6}$ & $0.946_{4}$ & $0.937_{3}$ & $0.909_{2}$ & \textbf{$0.873_{1}$} \\
DenseNet40 &  & $1.295_{12}$ & $1.197_{3}$ & $1.205_{5}$ & $1.273_{9}$ & $1.286_{11}$ & $1.285_{10}$ & $1.273_{8}$ & $1.208_{6}$ & $1.212_{7}$ & $1.202_{4}$ & $1.147_{2}$ & \textbf{$1.126_{1}$} \\
ResNet32W &  & $1.047_{12}$ & $0.954_{10}$ & $0.890_{4}$ & $0.935_{7}$ & $0.944_{8}$ & $0.966_{11}$ & $0.947_{9}$ & $0.878_{2}$ & $0.902_{5}$ & $0.925_{6}$ & $0.879_{3}$ & \textbf{$0.795_{1}$} \\ 
\hline
ResNet152SD & SVHN & $0.104_{8}$ & $0.100_{5}$ & $0.122_{11}$ & $0.089_{3}$ & $0.104_{7}$ & $0.230_{12}$ & $0.108_{9}$ & $0.109_{10}$ & $0.103_{6}$ & $0.099_{4}$ & $0.089_{2}$ & \textbf{$0.087_{1}$} \\ 
\hline
\multicolumn{2}{|l|}{\textbf{Average Rank }} & \multicolumn{1}{l}{10.73} & 5.27 & 7.82 & 6.55 & 9.09 & 10.82 & 8.09 & 5.73 & 5.64 & 4.27 & 2.64 & 1.36 \\ 
\hline
\multicolumn{14}{|c|}{\textbf{Accuracy \% }} \\ 
\hline
LeNet & \multirow{5}{*}{CIFAR10} & $70.00_{9}$ & $70.00_{9}$ & $70.00_{9}$ & $73.67_{6}$ & $69.92_{12}$ & $75.50_{3}$ & $73.39_{7}$ & $73.01_{8}$ & $73.99_{4}$ & $73.93_{5}$ & \textbf{$76.83_{1}$} & \textbf{$76.83_{1}$} \\
ResNet110 &  & $91.16_{9}$ & $91.16_{9}$ & $91.57_{7}$ & $91.82_{6}$ & $91.54_{8}$ & $91.94_{5}$ & $91.08_{12}$ & $91.13_{11}$ & $91.98_{4}$ & $92.05_{3}$ & \textbf{$92.20_{1}$} & \textbf{$92.20_{1}$} \\
ResNet110SD &  & $90.79_{7}$ & $90.79_{7}$ & $90.61_{9}$ & $91.24_{5}$ & $90.14_{11}$ & $88.36_{12}$ & $90.61_{9}$ & $91.02_{6}$ & \textbf{$91.61_{3}$} & $91.45_{4}$ & \textbf{$91.61_{1}$} & \textbf{$91.61_{1}$} \\
DenseNet40 &  & $90.16_{11}$ & $90.16_{11}$ & $90.59_{7}$ & $90.32_{10}$ & $90.66_{6}$ & \textbf{$91.81_{1}$} & $90.59_{7}$ & $91.27_{4}$ & $90.91_{5}$ & $90.57_{9}$ & $91.41_{2}$ & $91.41_{2}$ \\
ResNet32W &  & $95.13_{11}$ & $95.13_{11}$ & $95.63_{5}$ & $95.19_{9}$ & $95.57_{7}$ & $95.17_{10}$ & $95.63_{5}$ & $95.55_{8}$ & $95.69_{3}$ & $95.67_{4}$ & \textbf{$96.16_{1}$} & \textbf{$96.16_{1}$} \\ 
\hline
LeNet & \multirow{5}{*}{CIFAR100} & $39.01_{10}$ & $39.01_{10}$ & $41.71_{5}$ & $41.72_{4}$ & $41.76_{3}$ & $38.45_{12}$ & $41.03_{8}$ & $41.35_{6}$ & $41.21_{7}$ & $40.96_{9}$ & \textbf{$41.80_{1}$} & \textbf{$41.80_{1}$} \\
ResNet110 &  & $67.33_{11}$  & $67.33_{11}$  & $70.28_{4}$  & $69.02_{6}$  & $70.36_{3}$  & $68.03_{8}$  & $67.68_{10}$  & $67.92_{9}$  & $68.81_{7}$  & $69.52_{5}$  & \textbf{$70.42_{1}$}  & \textbf{$70.42_{1}$} \\
ResNet110SD &  & $71.23_{9}$  & $71.23_{9}$  & $69.63_{11}$  & $73.82_{4}$  & $69.53_{8}$  & $68.92_{12}$  & $73.47_{7}$  & $73.48_{6}$  & $74.03_{3}$  & $73.75_{5}$  & \textbf{$74.35_{1}$}  & \textbf{$74.35_{1}$} \\
DenseNet40 &  & $66.51_{6}$  & $66.51_{6}$  & $66.78_{4}$  & $66.51_{6}$  & $66.74_{5}$  & $66.25_{9}$  & $65.79_{11}$  & $65.41_{12}$  & $65.99_{10}$  & $66.89_{3}$  & \textbf{$67.23_{1}$}  & \textbf{$67.23_{1}$} \\
ResNet32W &  & \textbf{$78.61_{1}$}  & \textbf{$78.61_{1}$} & $77.31_{12}$  & $77.79_{8}$  & $77.35_{11}$  & $78.55_{3}$  & $77.50_{10}$  & $78.35_{6}$  & $78.19_{7}$  & $77.68_{9}$  & $78.53_{4}$  & $78.53_{4}$ \\ 
\hline
ResNet152SD & SVHN & $97.60_{7}$  & $97.60_{7}$  & $97.74_{3}$  & $97.39_{10}$  & $97.71_{4}$  & $97.38_{11}$  & $97.32_{12}$  & $97.45_{9}$  & $97.67_{5}$  & $97.66_{6}$  & \textbf{$97.99_{1}$}  & \textbf{$97.99_{1}$} \\ 
\hline
\multicolumn{2}{|l|}{\textbf{Average Rank }} & 8.27  & 8.27 & 6.91  & 6.73  & 7.09  & 7.82  & 8.91  & 7.73  & 5.27  & 5.64 & 1.36  & 1.36 \\ 
\hline
\multicolumn{14}{|c|}{\textbf{ECE }} \\ 
\hline
LeNet & \multirow{5}{*}{CIFAR10} & $0.014_{2}$ & \textbf{$0.012_{1}$} & $0.015_{3}$ & $0.020_{5}$ & $0.035_{7}$ & $0.068_{12}$ & $0.045_{8}$ & $0.017_{4}$ & $0.063_{11}$ & $0.052_{9}$ & $0.059_{10}$ & $0.021_{6}$ \\
ResNet110 &  & $0.041_{11}$ & \textbf{$0.015_{1}$} & $0.016_{2}$ & $0.031_{6}$ & $0.031_{7}$ & $0.071_{12}$ & $0.033_{8}$ & $0.017_{3}$ & $0.039_{10}$ & $0.030_{5}$ & $0.035_{9}$ & $0.026_{4}$ \\
ResNet110SD &  & $0.039_{11}$ & \textbf{$0.014_{1}$} & $0.033_{10}$ & $0.023_{2}$ & $0.032_{8}$ & $0.042_{12}$ & $0.024_{3}$ & $0.027_{4}$ & $0.029_{7}$ & $0.027_{5}$ & $0.032_{9}$ & $0.028_{6}$ \\
DenseNet40 &  & $0.043_{11}$ & $0.020_{4}$ & $0.021_{6}$ & $0.026_{10}$ & $0.022_{7}$ & $0.050_{12}$ & $0.024_{8}$ & $0.020_{5}$ & $0.018_{3}$ & $0.025_{9}$ & $0.016_{2}$ & \textbf{$0.016_{1}$} \\
ResNet32W &  & $0.027_{11}$ & $0.016_{4}$ & $0.024_{10}$ & $0.022_{8}$ & $0.023_{9}$ & $0.057_{12}$ & $0.016_{3}$ & $0.013_{1}$ & $0.019_{6}$ & \textbf{$0.014_{2}$} & $0.022_{7}$ & $0.016_{5}$ \\ 
\hline
LeNet & \multirow{5}{*}{CIFAR100} & $0.032_{6}$ & \textbf{$0.015_{3}$} & $0.025_{5}$ & $0.035_{7}$ & $0.039_{10}$ & $0.018_{4}$ & $0.035_{8}$ & \textbf{$0.015_{2}$} & $0.042_{11}$ & $0.036_{9}$ & $0.074_{12}$ & \textbf{$0.015_{1}$} \\
ResNet110 &  & $0.133_{12}$ & $0.030_{3}$ & $0.056_{9}$ & $0.037_{5}$ & $0.064_{11}$ & $0.022_{2}$ & $0.036_{4}$ & $0.047_{8}$ & $0.043_{6}$ & \textbf{$0.019_{1}$} & $0.057_{10}$ & $0.046_{7}$ \\
ResNet110SD &  & $0.087_{12}$ & \textbf{$0.015_{1}$} & $0.048_{9}$ & $0.016_{2}$ & $0.053_{10}$ & $0.024_{7}$ & $0.016_{3}$ & $0.035_{8}$ & $0.020_{4}$ & $0.023_{5}$ & $0.054_{11}$ & $0.024_{6}$ \\
DenseNet40 &  & $0.105_{12}$ & $0.031_{6}$ & $0.045_{9}$ & \textbf{$0.022_{2}$} & $0.050_{11}$ & $0.040_{8}$ & $0.024_{3}$ & $0.027_{4}$ & $0.028_{5}$ & $0.034_{7}$ & $0.068_{10}$ & $0.035_{1}$ \\
ResNet32W &  & $0.119_{12}$ & $0.097_{11}$ & $0.066_{9}$ & $0.042_{3}$ & $0.076_{10}$ & \textbf{$0.022_{1}$} & $0.048_{5}$ & $0.064_{8}$ & $0.058_{7}$ & $0.045_{4}$ & $0.049_{6}$ & $0.041_{2}$ \\ 
\cline{1-14}
ResNet152SD & SVHN & $0.009_{7}$ & $0.007_{5}$ & $0.031_{10}$ & $0.006_{3}$ & $0.033_{11}$ & $0.061_{12}$ & $0.005_{1}$ & $0.010_{8}$ & $0.006_{4}$ & \textbf{$0.005_{1}$} & $0.015_{9}$ & $0.008_{6}$ \\ 
\hline
\multicolumn{2}{|l|}{\textbf{Average Rank }} & 9.73 & 3.64 & 7.45 & 4.82 & 9.18 & 8.55 & 4.91 & 5.00 & 6.73 & 5.18 & 8.64 & 4.09 \\ 
\hline
\multicolumn{14}{|c|}{\textbf{cwECE }} \\ 
\hline
LeNet & \multirow{5}{*}{CIFAR10} & \textbf{$0.049_{2}$} & $0.050_{3}$ & $0.052_{5}$ & $0.060_{7}$ & $0.056_{6}$ & $0.073_{2}$ & $0.072_{9}$ & $0.047_{1}$ & $0.072_{8}$ & $0.073_{10}$ & $0.088_{12}$ & $0.051_{4}$ \\
ResNet110 &  & $0.153_{12}$ & $0.119_{8}$ & $0.023_{2}$ & $0.110_{7}$ & \textbf{$0.018_{1}$} & $0.085_{7}$ & $0.148_{10}$ & $0.031_{3}$ & $0.153_{11}$ & $0.144_{9}$ & $0.084_{5}$ & $0.037_{4}$ \\
ResNet110SD &  & $0.042_{10}$ & $0.084_{12}$ & $0.038_{7}$ & $0.032_{3}$ & $0.041_{9}$ & $0.050_{2}$ & $0.034_{6}$ & $0.033_{5}$ & $0.038_{8}$ & $0.033_{4}$ & $0.031_{2}$ & \textbf{$0.029_{1}$} \\
DenseNet40 &  & $0.047_{11}$ & \textbf{$0.027_{3}$} & $0.033_{9}$ & $0.041_{10}$ & $0.029_{4}$ & $0.054_{1}$ & $0.033_{7}$ & $0.030_{6}$ & \textbf{$0.027_{2}$} & $0.033_{7}$ & $0.030_{5}$ & \textbf{$0.027_{1}$} \\
ResNet32W &  & $0.030_{11}$ & $0.019_{4}$ & $0.027_{10}$ & $0.022_{8}$ & $0.020_{6}$ & $0.063_{1}$ & $0.023_{9}$ & $0.020_{5}$ & $0.019_{3}$ & $0.021_{7}$ & \textbf{$0.015_{1}$} & \textbf{$0.015_{2}$} \\ 
\hline
LeNet & \multirow{5}{*}{CIFAR100} & $0.130_{4}$ & \textbf{$0.127_{2}$} & $0.132_{8}$ & $0.132_{6}$ & $0.131_{5}$ & $0.128_{10}$ & $0.135_{9}$ & $0.132_{7}$ & $0.137_{10}$ & $0.125_{1}$ & $0.140_{11}$ & $0.142_{12}$ \\
ResNet110 &  & $0.145_{12}$ & $0.105_{2}$ & \textbf{$0.104_{1}$} & $0.109_{5}$ & $0.114_{7}$ & $0.105_{10}$ & $0.130_{9}$ & $0.109_{6}$ & $0.129_{8}$ & $0.132_{10}$ & $0.141_{11}$ & $0.109_{4}$ \\
ResNet110SD &  & $0.107_{8}$ & $0.094_{2}$ & $0.104_{5}$ & $0.099_{3}$ & $0.104_{7}$ & $0.111_{4}$ & $0.117_{12}$ & $0.102_{4}$ & $0.104_{6}$ & $0.112_{11}$ & $0.112_{10}$ & \textbf{$0.093_{1}$} \\
DenseNet40 &  & $0.125_{8}$ & $0.113_{2}$ & $0.118_{6}$ & $0.114_{3}$ & $0.128_{10}$ & $0.137_{1}$ & $0.117_{5}$ & $0.124_{7}$ & $0.129_{11}$ & $0.127_{9}$ & $0.116_{4}$ & \textbf{$0.112_{1}$} \\
ResNet32W &  & $0.126_{12}$ & $0.116_{11}$ & $0.103_{3}$ & $0.109_{7}$ & $0.111_{9}$ & $0.108_{7}$ & $0.110_{8}$ & $0.094_{2}$ & $0.111_{10}$ & $0.108_{5}$ & $0.104_{4}$ & \textbf{$0.093_{1}$} \\ 
\hline
ResNet152SD & SVHN & $0.012_{7}$ & $0.011_{5}$ & $0.032_{10}$ & $0.010_{4}$ & $0.039_{11}$ & $0.138_{1}$ & $0.011_{6}$ & $0.012_{8}$ & $0.016_{9}$ & $0.009_{2}$ & \textbf{$0.008_{1}$} & $0.010_{3}$ \\ 
\hline
\multicolumn{2}{|l|}{\textbf{Average Rank }} & 8.82 & 4.91 & 6.00 & 5.73 & 6.82 & 4.18 & 8.18 & 4.91 & 7.82 & 6.82 & 6.00 & 3.09 \\
\hline
\end{tabular}}
\end{table*}

\section{Experimental Results}
\label{sec:exp}
The main goals of our experiments are to compare the generalization and calibration performance of our proposed methods ILS1, ILS2 and ILS to the existing methods over real-world image datasets.
We compare against the standard label smoothing (LS), one-level self-distillation (Dis) \cite{kim2020self}, class-based label smoothing (\emph{cLS}) \cite{liu2020class}, the bootstrap-soft approach (\emph{BS\_Soft}) \cite{reed2014training}, and the beta smoothing (\emph{Beta}) \cite{zhang2020self}.
For \emph{cLS}, we used the Euclidean distance as the similarity measure between instances (other methods proposed in \cite{liu2020class} are domain-specific). 
For \emph{BS\_Soft}, we used $\beta = 0.95$ according to the author's recommendation, representing $95\%$ of the hard target labels and $5\%$ of the network predictions.
For \emph{Beta}, we followed the author's recommendation by using $\alpha$ = $0.4$ and the parameter $a$ of the Beta distribution is set such that $\mathbb{E}[\alpha+(1-\alpha)b_{i}]=\epsilon_{LS}$ where $b_{i}$ is randomly sampled from $Beta(a,1)$, and $\epsilon_{LS}$ is the best smoothing factor found using Standard LS (LS).
Source code will be made available when published
\footnote{https://github.com/mmaher22/Instance-based-label-smoothing} 

\paragraph{Experimental Setup} We used 3 datasets (SVHN, CIFAR10 and CIFAR100) and 5 architectures (ResNet \cite{he2016deep}, Wide ResNet (ResNetW) \cite{zagoruyko2016wide}, DenseNet \cite{huang2017densely}, ResNet with stochastic depth \cite{huang2016deep}, and LeNet \cite{lecun1998gradient}), in total 11 different dataset-architecture pairs, all trained to minimize cross-entropy.
Following the same setup as used by Guo et al.~\cite{guo2017calibration}, 20\% of the training set was set aside for validation to tune the hyperparameters:
the smoothing factor $\epsilon\in\{0.01, 0.05, 0.1, 0.15, 0.2\}$ for LS and ILS2; parameters $\text{P1}\in\{0.975, 0.95, 0.925, 0.9, 0.85\}$ and $\text{P2}\in\{0.75, 1, 1.25, 1.5, 2\}$ for ILS1 and ILS; and the teacher's temperature when optimizing the student: $T\in\{1, 2, 4, 8, 16\}$ for ILS2 and ILS (in other uses of temperature scaling the temperature is optimized continuously with gradient descent as usual). The models from the epoch with the lowest validation loss are used in the evaluation and all experiments were repeated twice (different random seed), and the average result over the two replicates was reported. 
Further details of the setup are provided in the Appendix\textsuperscript{\ref{appendix_url}}.

\paragraph{Results} Table \ref{tab:results} summarizes the results, reporting cross-entropy, accuracy, ECE and cwECE for all 11 dataset-architecture pairs and providing the average ranks of all 11 methods across these.
Each of our proposed ILS-based methods (ILS1, ILS2, ILS, ILS+TempS) outperforms all existing methods in terms of the average rank according to both cross-entropy and accuracy.
We plotted the critical difference diagrams \cite{demvsar2006statistical} showing that the average rank of ILS+TempS is statistically significantly better than the average ranks of all existing methods both for cross-entropy and accuracy, being the best in all except two dataset-architecture pairs for accuracy and one dataset-architecture pair for cross-entropy (See the Appendix\textsuperscript{\ref{appendix_url}}).
In terms of calibration, the methods NoLS+TempS and ILS+TempS have better average ranks than all other methods in terms of ECE, and ILS+TempS and Beta are better in terms of cwECE but here some of the differences are not statistically different.
Among the best two, Beta is slightly better at ECE, and ILS+TempS is slightly better at cwECE.

The hyperparameters of ILS1 were tuned to improve cross-entropy and indeed, the average rank of ILS1 is better than LS in cross-entropy (but also in accuracy and cwECE, while losing in ECE).
The improvement of cwECE was the motivation of ILS2 but it was tuned on cross-entropy, whereas the results confirm that ILS2 is better than LS in cwECE, cross-entropy and accuracy, while losing in ECE. 
Combining ILS1 and ILS2 into ILS improves cross-entropy, accuracy and cwECE further, but harms ECE - this harm is undone by temperature scaling, as ILS+TempS gets the best average rank after NoLS+TempS.

Regarding the standard LS, the results on real data are mostly similar to the results on synthetic data (Table~\ref{tab:motivate-calibration}).
On synthetic data, the standard LS+TempS improved over NoLS+TempS in accuracy, cross-entropy and ECE while being worse in cwECE.
On real data, the same is true, except that LS+TempS now loses to NoLS+TempS in cross-entropy as well, probably because due to more classes the lack of classwise-calibration affects the cross-entropy more.
We provide results for ILS on the synthetic data too in the Appendix\textsuperscript{\ref{appendix_url}}.

\paragraph{Distillation Performance Gain}
It turns out that the knowledge distillation performance of teacher networks trained with ILS has been improved over No LS and Standard LS. Following previous work \cite{liexploring}, Table \ref{tab:distil} summarizes the performance of a student AlexNet \cite{krizhevsky2012imagenet} trained with knowledge distillation from different deeper teacher architectures (ResNet, DenseNet, Inception \cite{szegedy2017inception}) on CIFAR10-100, and Fashion-MNIST datasets. When the teacher model was trained without LS, the student network has a better accuracy compared to teacher trained with standard LS. However, ILS showed a consistent accuracy improvement over LS in all the given examples and slightly less than the case when the teacher was trained without LS. Additionally, the student network has better cross-entropy loss consistently when the teacher was trained with ILS.

\begin{table}
\centering
\caption{Distillation performance on a Student AlexNet given a teacher trained with NoLS, LS or ILS.}
\label{tab:distil}
\resizebox{0.45\textwidth}{!}{\begin{tabular}{|l|l|ccc|ccc|} 
\cline{3-8}
\multicolumn{1}{l}{} & \multicolumn{1}{c|}{} & \multicolumn{3}{c|}{\textbf{cross-entropy } } & \multicolumn{3}{c|}{\textbf{Accuracy \% } } \\ 
\hline
\textbf{Teacher} & \textbf{Dataset}  & \multicolumn{1}{c|}{\textbf{No LS }} & \multicolumn{1}{c|}{\textbf{LS} } & \textbf{ILS}  & \multicolumn{1}{l|}{\textbf{No LS }} & \multicolumn{1}{l|}{\textbf{LS} } & \multicolumn{1}{l|}{\textbf{ILS} } \\ 
\hline
ResNet18 & CIFAR10 & 0.972 & 0.951 & \textbf{0.937}  & 85.11 & 84.76 & \textbf{85.59}  \\
InceptionV4 & CIFAR10 & 0.999 & 0.964 & \textbf{0.912}  & \textbf{85.81}  & 84.61 & \textbf{85.81} \\
ResNet50 & CIFAR100 & 1.486 & 1.483 & \textbf{1.477}  & 60.58  & 60.42 & \textbf{60.59} \\
DenseNet40 & FashionMNIST & 0.725 & 0.722 & 0.716  & 93.50  & 93.50  & \textbf{93.53} \\
\hline
\end{tabular}}
\end{table}

\section{Conclusions}
\label{sec:conc}
Label smoothing has been adopted in many classification networks thanks to its regularization behaviour against noisy labels and reducing the overconfident predictions. In this paper, we showed empirically the drawbacks of standard label smoothing compared to hard-targets training using a synthetic dataset with known ground truth. Label smoothing was found to distort the similarity information among classes, which affects the model's class-wise calibration and distillation performance.

We proposed two new instance-based label smoothing approaches ILS1 and ILS2 motivated from network distillation.
Our approaches assign a different smoothing factor to each instance and distribute it non-uniformly based on the predictions of a teacher network trained on the same architecture without label smoothing. Our methods were evaluated on 11 different examples using 6 network architectures and 3 benchmark datasets. The experiments showed that ILS (Combining ILS1 and ILS2) gives a considerable and consistent improvement over the existing methods in terms of the cross-entropy loss, accuracy and calibration measured with confidence-ECE and classwise-ECE.

In ILS1, we have used a quadratic family of smoothing factor curves - research into other possible families and seeking theoretical justification for these curves remains as future work.

\section{Acknowledgment}
The Estonian Research Council supported this work under grant PUT1458.

\bibliographystyle{IEEEtran}
\bibliography{biblio}

\clearpage

\appendices

\section{Synthetic Datasets Experiments}
The synthetic datasets were used in our study to validate empirically the limitations of standard label smoothing. In these experiments, we sampled 100 datasets using different random seeds from a generative model. Table \ref{tab:appendix-synthetic2} shows the average results of these experiments including our proposed instance-based label smoothing methods over the 100 datasets. The ECE and cwECE metrics were calculated at 15 equal-width bins. We also present the average hyper-parameter values corresponding to each method. These parameters were tuned to get the best validation cross-entropy loss on each sampled dataset. It can be easily observed that our proposed methods ILS1, ILS2 and ILS are outperforming both No LS and Standard LS in all the performance metrics. ILS is slightly more overfitted to the validation set than ILS1 as it has one more hyper-parameter tuned over the validation set. For this reason, ILS has slightly lower accuracy, cross-entropy loss and ECE than ILS1. We believe that increasing the validation set size would even enhance more the performance of ILS.

\begin{table*}[!ht]
\centering
\caption{Comparison between the evaluated methods. Acc Dense/Sparse represent the accuracy on the test instances sampled from the dense/sparse regions separately. The Avg Hyper-parameter Values represent the average values of the different parameters corresponding to each method over the 100 sampled synthetic datasets.}
\label{tab:appendix-synthetic2}
\resizebox{\textwidth}{!}{
\begin{tabular}{|l|c|c|c|c|c|c|c|c|c|c|c|c|c|} 
\cline{5-14}
\multicolumn{1}{l}{} & \multicolumn{1}{c}{} & \multicolumn{1}{l}{} & \multicolumn{1}{l|}{} & \multicolumn{6}{c|}{\textbf{Test Performance}} & \multicolumn{4}{c|}{\textbf{Avg Hyper-parameter Values}} \\ 
\hline
\multicolumn{1}{|c|}{\textbf{Method} } & \begin{tabular}[c]{@{}c@{}}\textbf{Temp}\\\textbf{Scaling}\end{tabular} & \begin{tabular}[c]{@{}c@{}}\textbf{Avg}\\\textbf{Temperature}\end{tabular} & \begin{tabular}[c]{@{}c@{}}\textbf{Validation}\\\textbf{Loss}\end{tabular} & \textbf{Acc}  & \begin{tabular}[c]{@{}c@{}}\textbf{Acc}\\\textbf{Dense} \end{tabular} & \begin{tabular}[c]{@{}c@{}}\textbf{Acc}\\\textbf{Sparse} \end{tabular} & \begin{tabular}[c]{@{}c@{}}\textbf{Cross}\\\textbf{Entropy}\end{tabular} & \textbf{ECE}  & \textbf{cwECE}  & \begin{tabular}[c]{@{}c@{}}\textbf{Scaling}\\\textbf{Temp} \end{tabular} & \textbf{$\epsilon$}  & \textbf{$P_1$}~ & \textbf{$P_2$}~ \\ 
\hline
\textit{Bayes-Optimal }  & No & 1.000 & 0.4423 & \textit{81.98\% }  & 82.94\% & 78.14\% & \textit{0.4444}  & \textit{0.0067}  & \textit{0.0234}  & \textit{-}  & - & - & - \\ 
\hline
No LS & No & 1.000 & 0.5222 & 78.45\% & 81.43\% & 66.54\% & 0.5589 & 0.0426 & 0.0668 & - & - & - & - \\ 
\hline
No LS & Yes & 1.202 & 0.5092 & 78.45\% & 81.43\% & 66.54\% & 0.5446 & 0.0311 & 0.0632 & - & - & - & - \\ 
\hline
Standard LS & No & 1.000 & 0.4851 & 79.44\% & 81.17\% & 72.51\% & 0.5318 & 0.0293 & 0.0723 & - & 0.061 & - & - \\ 
\hline
Standard LS & Yes & 0.946 & 0.4797 & 79.48\% & 81.24\% & 72.44\% & 0.5238 & 0.0263 & 0.0701 & - & 0.079 & - & - \\ 
\hline
Standard LS-0.2 & No & 1.000 & 0.5073 & 79.39\% & 81.11\% & 72.53\% & 0.5439 & 0.0650 & 0.0785 & - & 0.200 & - & - \\ 
\hline
Standard LS-0.2 & Yes & 0.804 & 0.4788 & 79.39\% & 81.11\% & 72.53\% & 0.5301 & 0.0241 & 0.0707 & - & 0.200 & - & - \\ 
\hline
ILS1 & No & 1.000 & 0.4700 & \textbf{80.18\%} & 81.49\% & 74.93\% & \textbf{0.5041} & \textbf{0.0235} & 0.0495 & - & - &  &  \\ 
\hline
ILS1 & Yes & 0.942 & 0.4654 & 80.16\% & 81.48\% & 74.88\% & 0.5052 & 0.0268 & 0.0497 & - & - &  &  \\ 
\hline
ILS2 & No & 1.000 & 0.4804 & 79.56\% & 81.50\% & 73.37\% & 0.5232 & 0.0287 & 0.0496 & 11.48 & 0.0494 & - & - \\ 
\hline
ILS2 & Yes & 0.906 & 0.4744 & 79.58\% & \textbf{81.54\%} & 73.02\% & 0.5201 & 0.0259 & 0.0494 & 9.63 & 0.0698 & - & - \\ 
\hline
ILS & No & 1.000 & 0.4628 & 80.13\% & 81.42\% & \textbf{74.97\%} & 0.5089 & 0.0269 & 0.0449 & 2.91 & - & 1.63 & 0.80 \\ 
\hline
ILS & Yes & 1.033 & 0.4618 & 80.14\% & 81.45\% & 74.91\% & 0.5087 & 0.0278 & \textbf{0.0446} & 2.73 & - & 1.59 & 0.81 \\
\hline
\end{tabular}}
\end{table*}

\subsection{Dataset Generative model}
The generative model was shown in Figure \ref{fig:synthetic-dataset}, where the sum of the probability density functions of six Gaussian circles, 2-D Gaussian distributions with equal variances, of three classes, is plotted. Each two of these Gaussian circles are used to represent one class in the dataset. The classes are overlapped with different degrees to show that some classes could be more similar to each other. Additionally, one Gaussian circle for each class has 80\% of the probability density of this class generative model, representing more dense area of instances, while the other circle has only 20\% density, representing a sparse area of instances. This generative model is designed to ensure that neural networks would overfit the sparse area more than the dense area. Hence, using different smoothing factors is better for different feature space regions. Each training set includes 50 instances per class (150 instances in total). The validation set has the same size as the training set, while the testing set has 5000 instances per class. The instances of each class are randomly drawn out of the distribution of the two Gaussian circles of the corresponding class using SciPy multivariate normal sampler \footnote{\url{https://docs.scipy.org/doc/scipy/reference/generated/scipy.stats.multivariate\_normal.html}}. 

\subsection{Network Training}
A feed-forward neural network architecture was trained on each of the datasets in various modes (No LS, Standard LS, ILS) in addition to the Bayes-Optimal model computed given the generative model of the datasets. The network architecture consists of five hidden layers of 64 neurons each. ReLU activations were used in all the network layers. We used Adam optimizer with $10^{-2}$ learning rate. The learning rate was selected based on the learning rate finder \footnote{\url{https://github.com/davidtvs/pytorch-lr-finder}} results along a range of $10^{-7}$ to $1$. The networks were trained for a maximum of 500 epochs and an early stopping of 10 epochs. The networks from the epoch of the minimum validation loss were finally used in the evaluation.

\subsection{Hyper-parameter tuning}
We used the set of $\{0.001, 0.005, 0.01, 0.03, 0.05, ..., 0.19\}$ for the possible smoothing factor values ($\epsilon$) for LS and ILS2 methods. Also, the temperature value used with the teacher network final softmax in ILS2, and ILS was tuned from the set $\{1, 2, 4, 6\}$. In ILS1 and ILS, we set values of $\{0.75, 0.775, 0.8, 0.825, 0.85\}$ and $\{0.75, 1, 1.25, 1.5, 1.75, 2\}$ for the hyper-parameters $P_1$ and $P_2$, respectively. The best $\epsilon$ , $P_1$, $P_2$, and/or temperature values were selected for each dataset based on the best validation cross-entropy loss.

\subsection{Self-distillation Experiments}
In this set of experiments, we distilled a student network from a teacher of the same architecture but trained with different modes (NoLS, LS, ILS). The student was trained using the same optimizer and learning rate as the teacher network to minimize the log-loss with the teacher network predictions. The temperature values of both the student and teacher networks were set to $1$. The experiment was repeated for the 100 network replicas trained on the 100 sampled training sets as illustrated previously. The average results are reported in Table \ref{tab:motivate-distillation2} including the teacher trained with ILS.

\begin{table}[!ht]
\centering
\caption{Comparison between using teacher trained with/without LS in self-distillation.}
\label{tab:motivate-distillation2}
\resizebox{0.5\textwidth}{!}{
\begin{tabular}{|l|c|c|c|c|} 
\cline{2-5}
\multicolumn{1}{l|}{} & \multicolumn{4}{c|}{\textbf{Student Performance}} \\ 
\hline
\multicolumn{1}{|c|}{ \textbf{Teacher} } & \textbf{Accuracy}  & \textbf{CrossEntropy}  & \textbf{ECE}  & \textbf{cwECE}  \\ 
\hline
\textit{Bayes-Optimal}  & \textit{81.92\% }  & \textit{0.4451}  & \textit{0.0137}  & \textit{0.0338}  \\ 
\hline
No LS & 78.47\% & 0.5592  & \textbf{0.0405}  & \textbf{0.0649}  \\ 
\hline
Standard LS & 78.43\% & 0.5621 & 0.0408 & 0.0726 \\
\hline
ILS & \textbf{79.57}\% & \textbf{0.5201} & 0.0473 & 0.0661 \\
\hline
\end{tabular}}
\end{table}

\section{Smoothing Factor Value in ILS}
\label{sec:append_ils}
\begin{figure}[!ht]
  \begin{center}
    \includegraphics[width=0.4\textwidth]{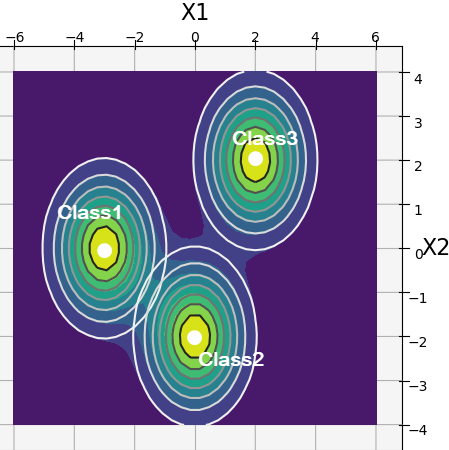}
  \end{center}
  \caption{Heat-map of the total density function of a generative model of 3-classes synthetic dataset with different separation between the classes.}
  \label{fig:synthetic2}
\end{figure}

\begin{figure*}[!ht]
  \begin{center}
    \includegraphics[width=\textwidth]{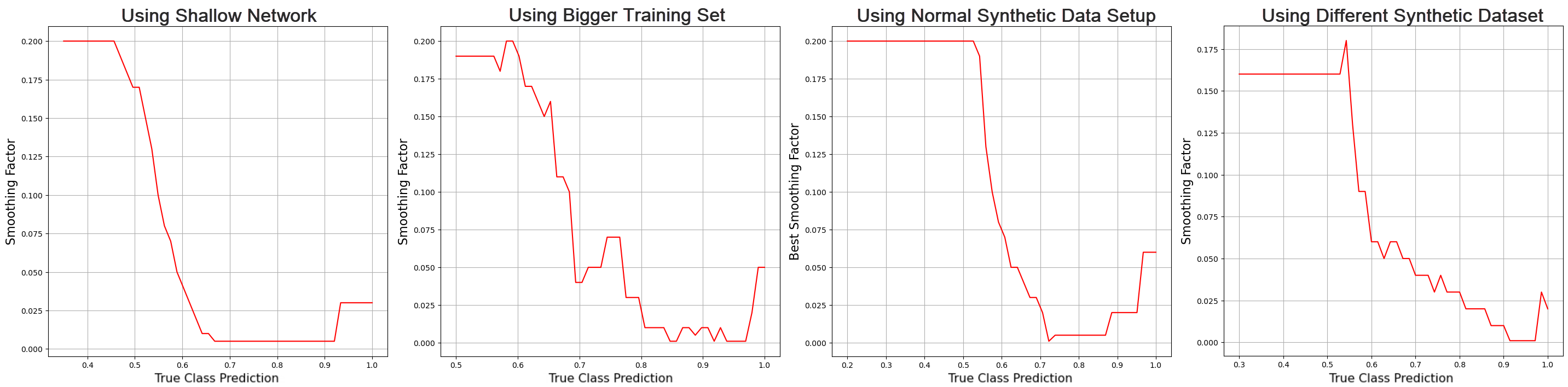}
  \end{center}
  \caption{Examples for the best smoothing factor curves corresponding to the predicted probability of the true class of the Bayes-Optimal model on different synthetic dataset setup.}
  \label{fig:eg-bo}
\end{figure*}

We generated multiple smoothing factor curves that can be used to decide the values of the smoothing factor corresponding to different true class predictions. As explained in Section \ref{sec:ils1}, the aim of these curves is to find the suitable amount of smoothing for each group of instances such that the network can predict similar true class probabilities as the Bayes-Optimal model. Figure \ref{fig:eg-bo} shows examples for the smoothing factor curves given the Bayes-Optimal predictions on synthetic datasets. To ensure that a similar behavior could be obtained with different experiment setup, we changed the neural network capacity, the training set size, and the separation between classes to observe the shape of these curves. First, we generated the smoothing factor curve on the same synthetic dataset experiment setup used in the main paper (Normal Synthetic Data Setup). Then, we tried out different variants of this setup like using a shallow network of a single hidden layer on the same generative synthetic dataset (Shallow Network), using larger training set of 200 instances per class instead of the originally used 50 instances (Bigger Training Set), and using different separation distances among the classes of the generative model of the dataset as shown in Figure \ref{fig:synthetic2} (Different Synthetic Dataset). We observed that the behavior of the obtained curves could be well approximated using a quadratic function of a shift parameter ($P_1$) on the x-axis and coefficient ($P_2$) representing the slope of the quadratic term.

\begin{figure*}[!ht]
  \begin{center}
    \includegraphics[width=\textwidth]{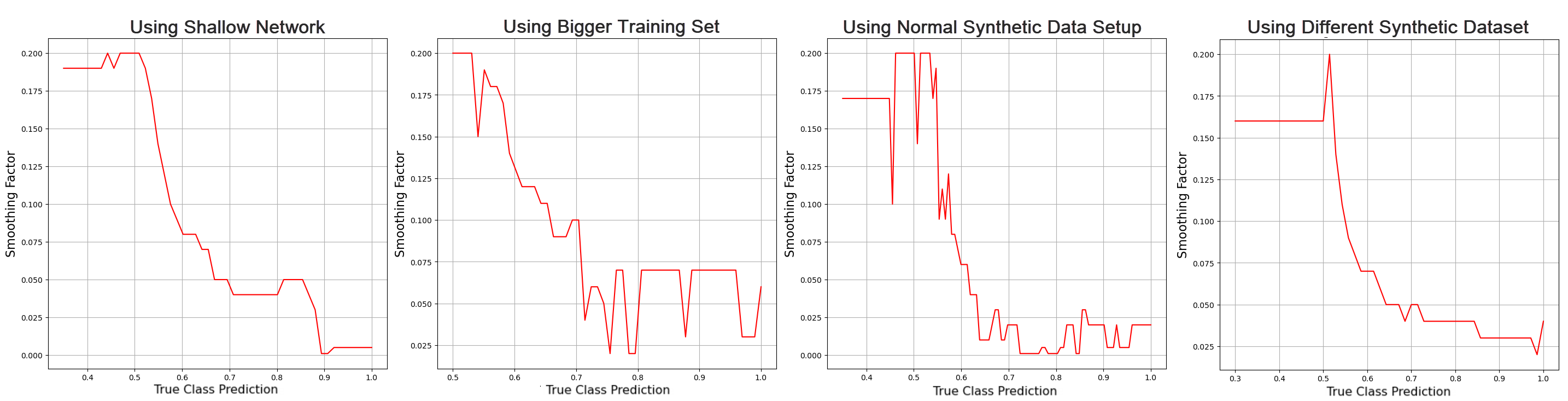}
  \end{center}
  \caption{Examples for the best smoothing factor curves corresponding to the predicted probability of the true class of the No-LS network + Temperature Scaling on different synthetic dataset setup.}
  \label{fig:eg-nols}
\end{figure*}

\begin{figure*}[!ht]
  \begin{center}
    \includegraphics[width=0.8\textwidth]{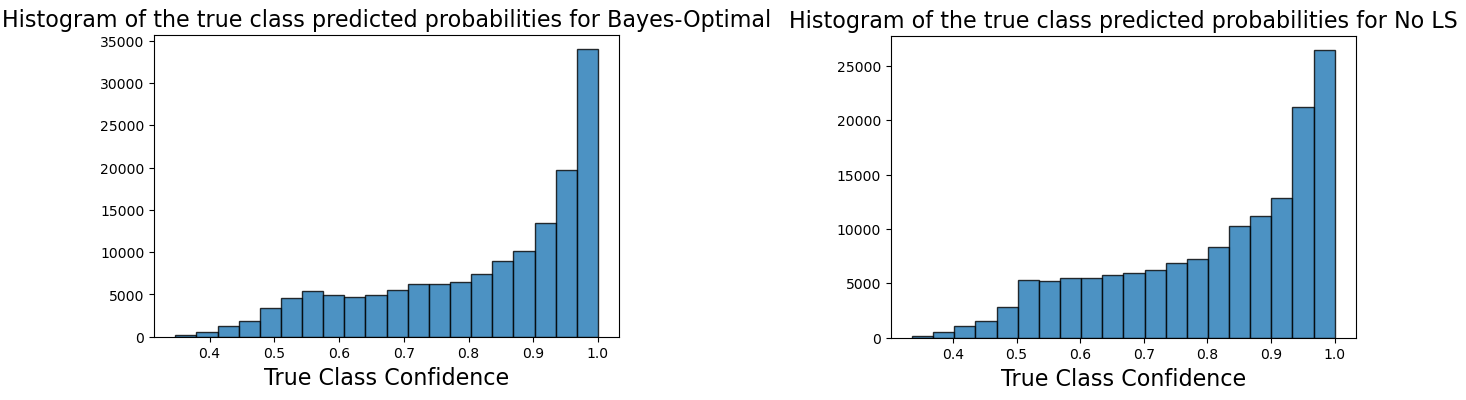}
  \end{center}
  \caption{Histograms of the true class predicted probabilities for BO, NoLS+TempS models on the normal synthetic dataset setup.}
  \label{fig:histbo}
\end{figure*}

\begin{figure*}[!ht]
  \begin{center}
    \includegraphics[width=\textwidth]{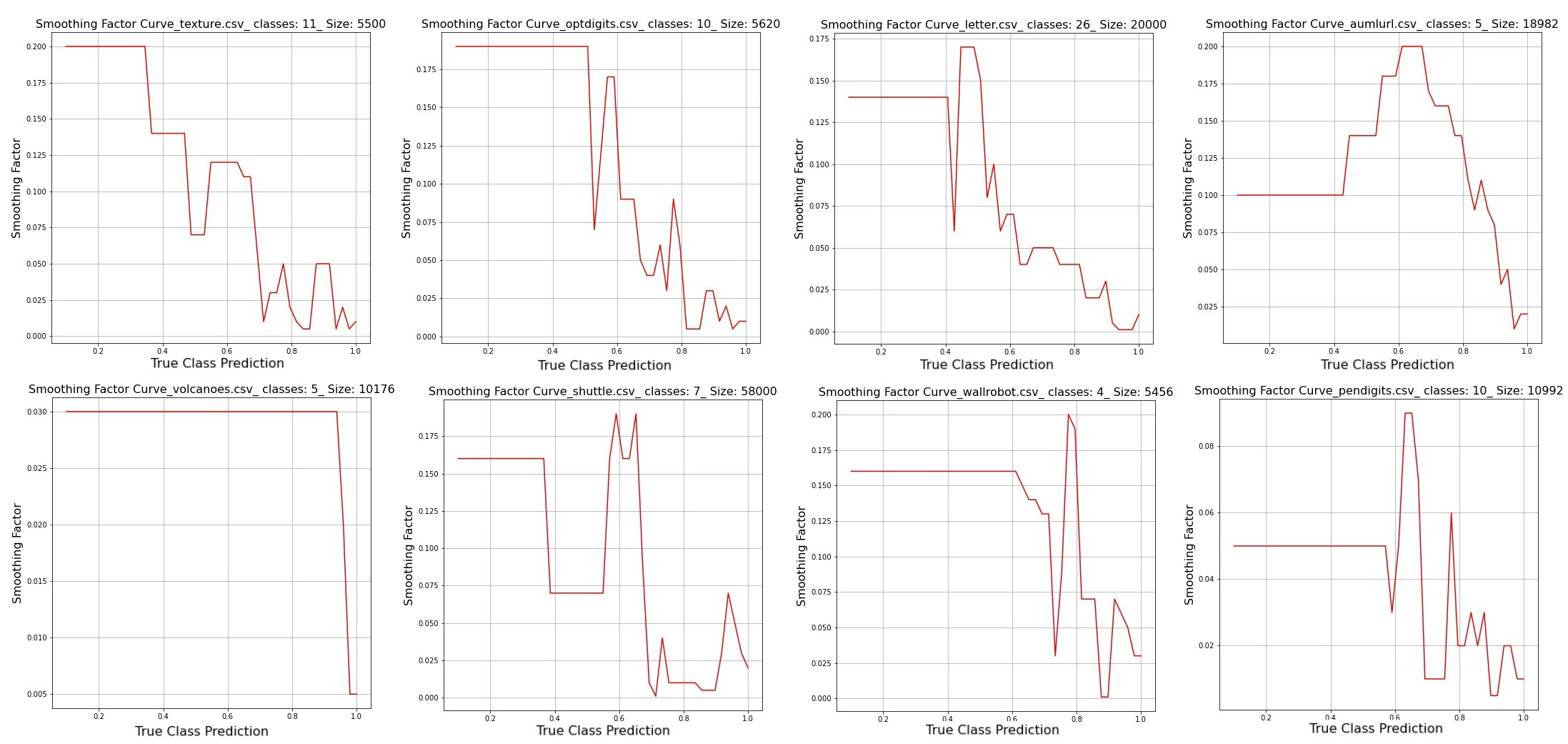}
  \end{center}
  \caption{Examples for the best smoothing factor curves corresponding to the predicted probability of the true class of the No-LS network + Temperature Scaling on different real datasets.}
  \label{fig:eg-real}
\end{figure*}

\begin{figure*}[!ht]
  \begin{center}
    \includegraphics[width=0.8\textwidth]{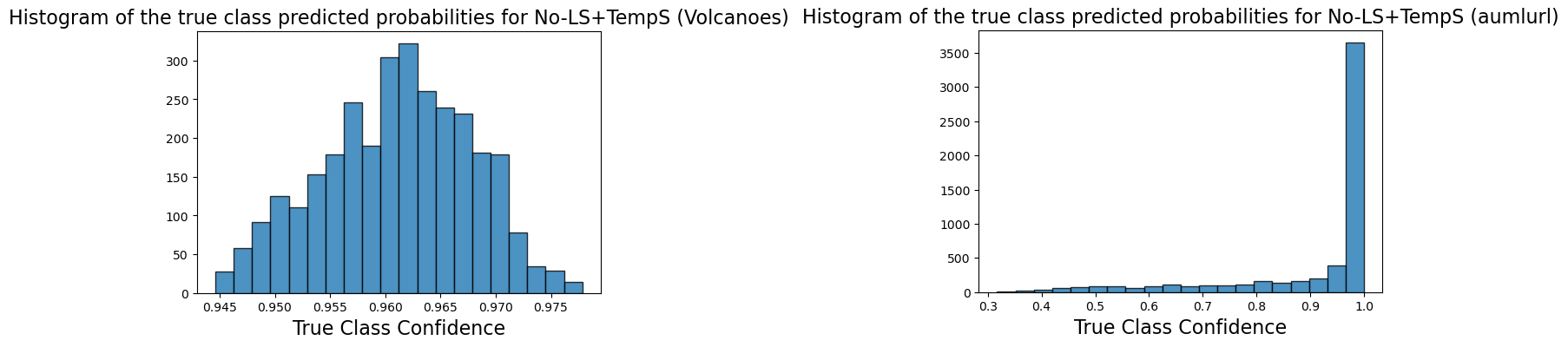}
  \end{center}
  \caption{Histograms of the true class predicted probabilities for BO, NoLS+TempS models on the volcanoes and amul-url datasets.}
  \label{fig:histreal}
\end{figure*}

Since knowing the Bayes-Optimal model is practically impossible, we also plotted the same smoothing factor curves given the No-LS with temperature scaling network predictions on the same synthetic dataset setup as shown in Figure \ref{fig:eg-nols}. Additionally, we generated the same curves on different multi-class real datasets with different number of instances, and classes obtained from OpenML \footnote{\url{https://www.openml.org/}}. A sample of these curves is shown in Figure \ref{fig:eg-real} and others could be found in our repository \footnote{https://github.com/mmaher22/Instance-based-label-smoothing}.
Interestingly, similar shapes for the smoothing factor curves to that shown in Figure \ref{fig:eg-bo} could be still observed. There could be some observable mismatches that can not be approximated using the quadratic family of functions especially on the curves of the real datasets. These mismatches occur in regions with very few instances and thus irrelevant in practice. These regions are mostly the left-side parts of the mapping curves (predicted probabilities less than 0.5). Figures \ref{fig:histbo} and \ref{fig:histreal} show histograms of predicted probabilities of the actual class on the normal synthetic dataset setup by the Bayes-Optimal and NoLS+TempS networks, and some real datasets by the NoLS+TempS network. It can be observed that the majority of the predictions are highly confident. For instance, in volcanoes dataset, all the true class probability predictions are $>0.94$. We choose the No-LS with temperature scaling network as it showed the best class-wise calibration performance. Hence, we will choose the instance-specific smoothing factor value based on the true class prediction of this network.

\begin{figure}[!ht]
  \begin{center}
    \includegraphics[width=0.5\textwidth]{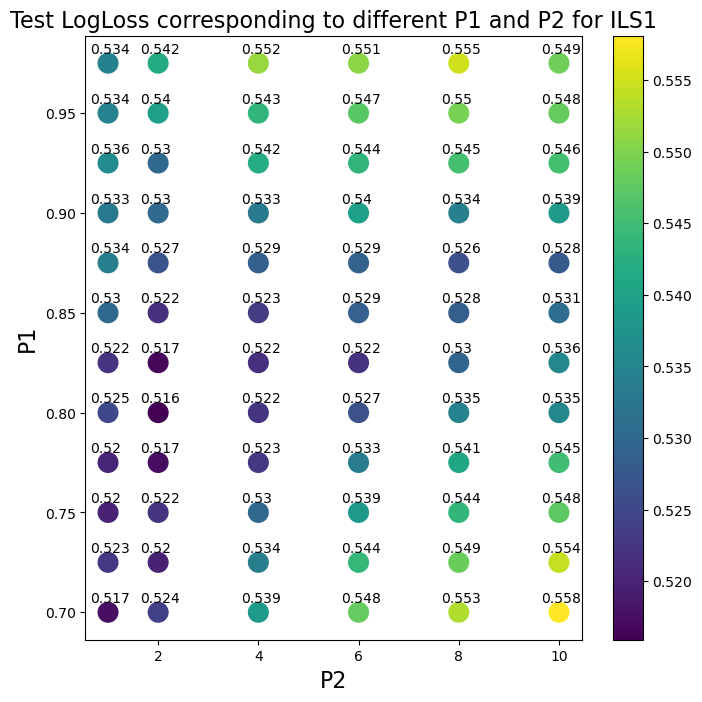}
  \end{center}
  \caption{The testing cross-entropy of ILS1 corresponding a grid of $P_1$ and $P_2$ parameters for the synthetic datasets.}
  \label{fig:scatter}
\end{figure}

To select good ranges for $P_1$ and $P_2$, we performed a grid search on a long range of values ranging from $0.7$ to $0.975$ for $P_1$, and from $1$ to $10$ for $P_2$ on the synthetic datasets. The summary of the this experiments is shown in Figure \ref{fig:scatter}, where the plot shows the average cross-entropy loss on the testing set for each combination of $P_1$ and $P_2$ in ILS1. This plot helped in narrowing down the promising ranges of $P_1$ and $P_2$ in our evaluation experiments on the synthetic data. Surprisingly, the best hyper-parameter settings ($P_1=0.8$, $P_2=2$), in Figure \ref{fig:scatter}, were found to be approximating the smoothing factor curve in Figure \ref{fig:ils1} to a very good extent. The scatter plot also confirms the good selection for the quadratic family of functions in Equation \ref{eq:ils1} to approximate the smoothing factor curves because changing the values of either $P_1$ or $P_2$ from the closest matching function parameters ($P_1=0.8$, $P_2=2$) leads to increase in the test loss. Hence, it would be easy to choose a suitable function out of this family by tuning the values of $P_1$ and $P_2$ on a separate validation set.

\section{Real Datasets Experiments}
\begin{figure}[!ht]
  \begin{center}
    \includegraphics[width=0.5\textwidth]{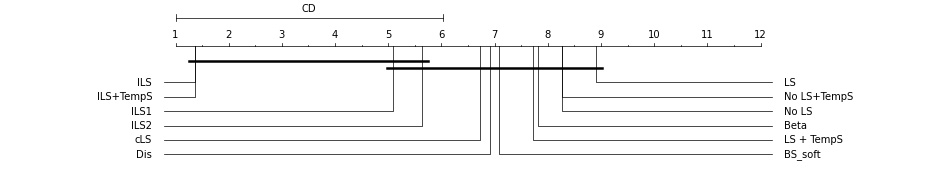}
  \end{center}
  \caption{The Critical Difference diagram of different methods on the real datasets in terms of the test accuracy.}
  \label{fig:test-acc}
\end{figure}

\begin{figure}[!ht]
  \begin{center}
    \includegraphics[width=0.5\textwidth]{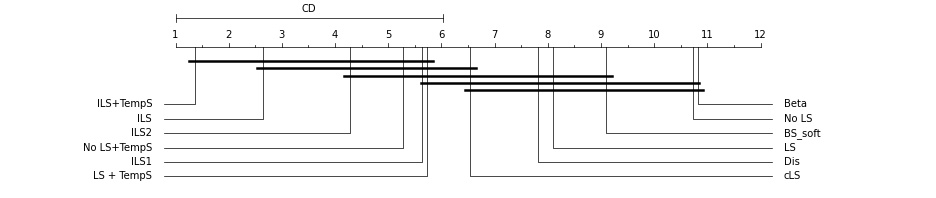}
  \end{center}
  \caption{The Critical Difference diagram of different methods on the real datasets in terms of the test log-loss.}
  \label{fig:test-loss}
\end{figure}

\begin{figure}[!ht]
  \begin{center}
    \includegraphics[width=0.5\textwidth]{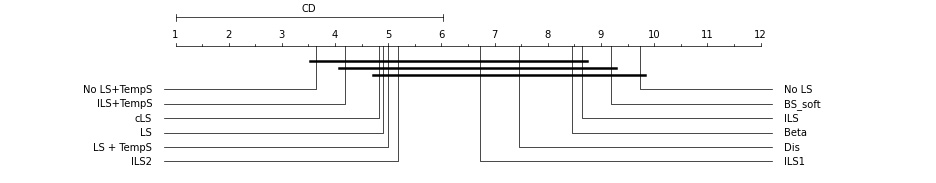}
  \end{center}
  \caption{The Critical Difference diagram of different methods on the real datasets in terms of the test ECE.}
  \label{fig:test-ece}
\end{figure}

\begin{figure}[!ht]
  \begin{center}
    \includegraphics[width=0.5\textwidth]{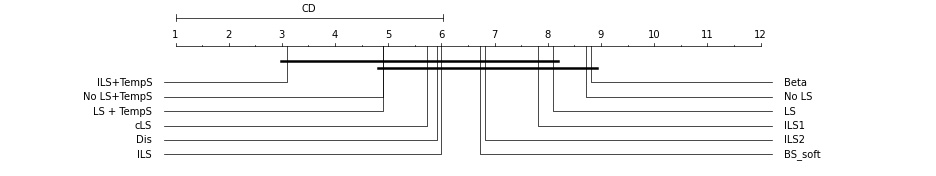}
  \end{center}
  \caption{The Critical Difference diagram of different methods on the real datasets in terms of the test cwECE.}
  \label{fig:test-cwece}
\end{figure}

In this set of experiments, we used the CIFAR10/100 \footnote{\url{https://www.cs.toronto.edu/~kriz/cifar.html}} and SVHN \footnote{\url{http://ufldl.stanford.edu/housenumbers/}} datasets. For all datasets, we used the original training and testing splits of 50K and 10K images respectively for CIFAR10/100, and 604K and 26K images for SVHN dataset. The training splits were further divided into (45K-5K) images and (598K-6K) images as training and validation splits for CIFAR10/100 and SVHN, respectively. For data augmentation, we normalized both CIFAR10/100 and SVHN datasets and we used horizontal flips and random crops for the training splits of CIFAR10/100. The experiments were repeated two times with different seeds for the whole training process including the training/validation splitting, and the average results are reported. The critical difference diagrams for Nemenyi two tailed test given significance level $\alpha = 5\%$ for the average ranking of the evaluated methods and the number of tested datasets in terms of test accuracy, cross-entropy loss, ECE and cwECE are summarized in Figures \ref{fig:test-acc},\ref{fig:test-loss}, \ref{fig:test-ece} and \ref{fig:test-cwece}, respectively.

\subsection{Architectures}
In these experiments, we trained the networks following their original papers as described below. The smoothing factor $\epsilon$ and the temperature scaling factor of the teacher network in the proposed method ILS2 were tuned from the sets \{0.01, 0.05, 0.1, 0.15, 0.2\} and \{1, 2, 4, 8, 16\}, respectively. Also, the hyper-parameters $P_1$ and $P_2$ in ILS1 were tuned from the sets $\{0.975, 0.95, 0.925, 0.9, 0.85\}$ and $\{0.75, 1, 1.25, 1.5, 2\}$, respectively. We selected the epoch with the best validation loss to report the results.

\paragraph{LeNet}
The LeNet model was used in the CIFAR10/100 experiments. We used the open source implementation of LeNet available in \footnote{\url{https://github.com/meliketoy/wide-resnet.pytorch/blob/master/networks/lenet.py}}. The network was trained for $300$ epochs using the stochastic gradient descent (SGD) optimizer with $0.9$ momentum and an initial learning rate of $0.1$ that is reduced by a pre-defined scheduler with a rate of $10\%$ at the $60^{th}$, $120^{th}$, and $160^{th}$ epochs. The mini-batch size and the weight decay were set to $128$ and $0.0001$, respectively.

\paragraph{DenseNet}
The DenseNet model was used with the general CIFAR10/100 datasets experiments, and knowledge distillation experiments on Fashion-MNIST. We used the open source implementation of ResNet available in \footnote{\url{https://github.com/pytorch/vision/blob/master/torchvision/models/densenet.py}}. The network was trained for $300$ epochs using the stochastic gradient descent (SGD) optimizer with $0.9$ momentum and an initial learning rate of $0.1$ that is reduced by a pre-defined scheduler with a rate of $10\%$ at the $150^{th}$ and $225^{th}$ epochs. The mini-batch size and the weight decay were set to $64$ and $0.0001$, respectively.

\paragraph{ResNet}
The ResNet model was used with the general CIFAR10/100 datasets experiments, and knowledge distillation experiments. We used the open source implementation of ResNet available in \footnote{\url{https://github.com/pytorch/vision/blob/master/torchvision/models/resnet.py}}. The network was trained for $200$ epochs using the stochastic gradient descent (SGD) optimizer with $0.9$ momentum and an initial learning rate of $0.1$ that is reduced by a pre-defined scheduler with a rate of $10\%$ at the $80^{th}$ and $150^{th}$ epochs. The mini-batch size and the weight decay were set to $128$ and $0.0001$, respectively.

\paragraph{ResNet with Stochastic Depth}
The ResNet with Stochastic Depth model (ResNetSD) was used in the CIFAR10/100 and SVHN experiments. We used the open source implementation of ResNetSD available in \footnote{\url{https://github.com/shamangary/Pytorch-Stochastic-Depth-Resnet}}. The network was trained for $500$ epochs using the stochastic gradient descent (SGD) optimizer with $0.9$ momentum and an initial learning rate of $0.1$ that is reduced by a pre-defined scheduler with a rate of $10\%$ at the $250^{th}$ and $325^{th}$ epochs. The mini-batch size and the weight decay were set to $128$ and $0.0001$, respectively.

\paragraph{Wide ResNet}
The Wide ResNet model (ResNetW) was used in the CIFAR10/100 experiments. We used the open source implementation of ResNetW available in \footnote{\url{https://github.com/meliketoy/wide-resnet.pytorch}}. The network was trained for $200$ epochs using the stochastic gradient descent (SGD) optimizer with $0.9$ momentum and an initial learning rate of $0.1$ that is reduced by a pre-defined scheduler with a rate of $20\%$ at the $60^{th}$, $120^{th}$, and $160^{th}$ epochs. The mini-batch size, growth rate and the weight decay were set to $128$, $10$ and $0.0005$, respectively.

\paragraph{InceptionV4}
The Inception model was used with the knowledge distillation experiment on CIFAR10 dataset. We used the open source implementation of InceptionV4 available in \footnote{\url{https://github.com/weiaicunzai/pytorch-cifar100}}. The network was trained for $200$ epochs using the stochastic gradient descent (SGD) optimizer with $0.9$ momentum and an initial learning rate of $0.1$ that is reduced by a pre-defined scheduler with a rate of $10\%$ at the $30^{th}$, $60^{th}$, and $90^{th}$ epochs. The mini-batch size and the weight decay were set to $256$ and $0.0001$, respectively.

\subsection{Knowledge Distillation Experiments}
In these experiments, we distilled multiple student AlexNets from different teacher architectures trained with/without LS or with our proposed ILS. We used the default training and testing splits of CIFAR10/100 and Fashion-MNIST datasets. The experiments were repeated 3 times with different random seeds and the average results are reported. We used the open-source implementation available in \footnote{\url{https://github.com/peterliht/knowledge-distillation-pytorch}}. During distillation, we trained the student network by minimizing a weighted average of the log-loss between the student and the teacher outputs, and the log-loss between the student and the hard target labels with ratios of $0.9$ and $0.1$, respectively. using Adam optimizer. We set the initial learning rate to $10^{-2}$ and the mini-batch size to $256$. The network was trained for $150$ epochs and the learning rate was reduced using a predefined scheduler by $10\%$ after each $150$ iterations. The temperature values were set to $20$ and $1$ for the teacher and student networks, respectively.

\end{document}